\pdfoutput=1
\documentclass[11pt]{article}

\usepackage{authblk}
\usepackage{EMNLP2023}

\usepackage{times}
\usepackage{latexsym}
\usepackage[T1]{fontenc}
\usepackage[utf8]{inputenc}

\usepackage{microtype}

\usepackage{inconsolata}

\usepackage{amsmath,amssymb,graphicx}
\usepackage{tabularx}
\usepackage{multirow}
\usepackage{ctable,booktabs}
\usepackage[inline]{enumitem}
\usepackage{makecell}
\usepackage{subcaption}

\usepackage[linesnumbered, ruled, ruled, noend]{algorithm2e}
\usepackage{setspace}
\usepackage{etoolbox}
\usepackage{xspace}

\makeatletter
\patchcmd{\@algocf@start}% <cmd>
  {-1.5em}% <search>
  {0pt}% <replace>
  {}{}% <success><failure>
\makeatother

\SetKwInOut{Input}{Input}  
\newlength\mylen

\graphicspath{{images/}}
\newcolumntype{P}[1]{>{\centering\arraybackslash}p{#1}}

\newcommand{\mcv}{{\textsc{mcv\_accent}}\xspace}

\title{Accented Speech Recognition With Accent-specific Codebooks}

\newcommand*{\emails}{%
   \normalsize \texttt{\{darshanp,pjyothi,vinit\}@cse.iitb.ac.in$^{\ddagger}$, sriramg@iisc.ac.in$^{\S}$}
}

\author[$\ddagger$]{\textbf{Darshan Prabhu}}
\author[$\ddagger$]{\textbf{Preethi Jyothi}}
\author[$\S$]{\textbf{Sriram Ganapathy}}
\author[$\ddagger$]{\textbf{Vinit Unni}}
\affil[$\ddagger$]{Indian Institute of Technology Bombay, Mumbai, India}
\affil[$\S$]{Indian Institute of Science, Bangalore, India}
\affil[ ]{ \vspace{-0.25cm} }
\affil[ ]{ \emails }
\begin{document}
\maketitle
\begin{abstract}
Speech accents pose a significant challenge to state-of-the-art automatic speech recognition (ASR) systems. Degradation in performance across underrepresented accents is a severe deterrent to the inclusive adoption of ASR. In this work, we propose a novel accent adaptation approach for end-to-end ASR systems using cross-attention with a trainable set of codebooks. These learnable codebooks capture accent-specific information and are integrated within the ASR encoder layers. The model is trained on accented English  speech, while the test data also contained accents which were not seen during training.  On the Mozilla Common Voice multi-accented dataset, we show that our proposed approach yields significant performance gains not only on the seen English accents (up to $37\%$ relative improvement in word error rate) but also on the  unseen accents (up to $5\%$ relative improvement in WER). Further, we  illustrate benefits for a zero-shot transfer setup on the L2Artic dataset. We also compare the performance with other approaches based on accent adversarial training.

%\def\thefootnote{}\footnotetext{Our data splits and codebase are available at \href{https://github.com/csalt-research/accented-codebooks-asr}{https://github.com/csalt-research/accented-codebooks-asr}.}\def\thefootnote{\arabic{footnote}}

%Accent in speech pose a significant challenge for Automatic Speech Recognition (ASR) systems. The degradation in performance across accents is a severe deterrent to the inclusive adoption of ASR. In this work, we propose a novel ASR system that introduces accent specific and accent generic codebooks into an end-to-end ASR architecture. These learnable codebooks are trained to learn nuances of a specific accent and are injected into the encoder layers of an end-to-end Conformer ASR model using cross-attention. Cross-attention enables a more fine-grained integration of codebooks into the ASR system resulting in them learning accent specific traits. We experiment with 5 seen and 11 unseen accents from the Mozilla Common Voice dataset. We show word error rate reductions across all accents with around  1\% absolute reduction on unseen accents. We outperform popular accent-adaptation techniques, including domain adversarial training and I-vector based approaches.
\end{abstract}

\section{Introduction}
\label{sec:intro}

Accents in speech typically refer to the distinctive way in which the words are pronounced by diverse speakers. While a speaker's accent may be primarily derived from their native language, speech accents are also influenced by various other factors related to the geographic location, educational background,  socio-economic and socio-linguistic factors like race, gender and cultural  diversity~\cite{benzeghiba2007automatic}.  It is therefore infeasible to build automatic speech recognition (ASR) systems which comprehensively cover speech accents during training. In such scenarios, novel speech accents continue to have an adverse effect on ASR performance~\cite{Beringer1998GermanRV, aksenova2022accented}. While humans effectively recognize speech from new and unseen accents~\cite{clarke2004rapid}, ASR systems show substantial degradation in performance when dealing with new accents that are unseen during training~\cite{chu2021accented}.

Prior works attempting to address accent-related challenges for ASR can be categorized into three groups: i) multi-accent training~\cite{huang2014multi, 7846328}, ii) accent-aware training using accent embeddings~\cite{jain2018improved} or adversarial learning~\cite{dat}, and iii) accent adaptation using supervised~\cite{7953071, winata2020learning} or unsupervised techniques~\cite{turan2020achieving}. While partial success has been achieved using most of these approaches, the development of robust speech recognition systems that are invariant to accent differences in training and test remains a challenging problem.

In this work, we propose a new codebook based technique for accent adaptation of state-of-the-art Conformer-based end-to-end (E2E) ASR models~\cite{conformer}. For each of the accents observed in the training data, we define a codebook with a predefined number of randomly-initialized vectors. These accent codes are integrated with the self-attended representations in each encoder layer via the cross-attention mechanism, similar to the perceiver framework \cite{jaegle2021perceiver}. The ASR model is trained on multi-accented data with standard end-to-end (E2E) ASR objectives. The codes capture accent-specific information as the training progresses. During inference, we propose a beam-search decoding algorithm that searches over a combined set of hypotheses obtained by using each set of accent-specific codes (once for each seen accent) with the trained ASR model. On the Mozilla Common Voice (MCV) corpus, we observe significant improvements on both seen and new accents at test-time compared to the baseline and existing supervised accent-adaptation techniques.

Our main contributions are:
\begin{itemize}
\item We propose a new accent adaptation technique for Conformer-based end-to-end ASR models using cross-attention over a set of learnable codebooks. Our technique comprises learning accent-specific codes during training and a new beam-search decoding algorithm to perform an optimized combination of the codes from the seen accents. We demonstrate significant performance improvements on both seen and unseen accents over competitive baselines on the MCV dataset.
\item Even on a zero-shot setting involving a new accented evaluation set, L2-Arctic~\cite{l2arctic}, we show significant improvements using our codebooks trained using MCV.
\item We publicly release our train/development/test splits spanning different seen and unseen accents in the MCV corpus. Reproducible splits on MCV have been entirely missing in prior work and we hope this will facilitate fair comparisons across existing and new accent-adaptation techniques.%
\footnote{The MCV data splits and codebase are available at: \href{https://github.com/csalt-research/accented-codebooks-asr}{https://github.com/csalt-research/accented-codebooks-asr}.}
%
% Code to replicate our experiments is also planned for release.
\end{itemize}

\section{Related Work}
\label{sec:rel_work}
Traditional cascaded ASR systems~\cite{tradASR} handled accents by either modifying the pronunciation dictionary~\cite{tradphone1,tradphone2} or modifying the acoustic model~\cite{tradmodel1,tradmodel2}. More recent work on accented ASR has focused on building end-to-end accent-robust ASR models. Towards this, there are two sets of prior works: \emph{Accent-agnostic} approaches and \emph{Accent-aware} approaches. 

\paragraph{Accent-agnostic ASR.} Such approaches force the model to disregard the accent information present in the speech and focus only on the underlying content. Prior work based on this approach uses adversarial training~\cite{domadv} or similarity losses. Using domain adversarial training, with the discriminator being an accent classifier, has shown significant improvements over standard ASR models~\cite{dat}. Pre-training the accent classifier~\cite{bobw} and clustering-based accent relabelling~\cite{redat} have also led to further performance improvements. The use of generative adversarial networks for this task has also been explored~\cite{gat}. Rather than being explicitly domain adversarial, other accent agnostic approaches use cosine losses~\cite{contr1} or contrastive losses~\cite{contrloss,contr2} to make the model accent neutral. These losses force the model to output similar representations for inputs with the same underlying transcript. 

\paragraph{Accent-aware ASR.} Accent-aware approaches feed the model additional information about the accent of the input speech. Early work in this category focused on using the multi-task learning (MTL) paradigm~\cite{mtl1,jain2018improved,mtl2} that jointly trains accent-specific auxiliary tasks with ASR. Different types of embeddings like i-vectors~\cite{saon2013speaker, ivec}, dialect symbols~\cite{li2017multidialect}, embeddings extracted from TDNN models~\cite{jain2018improved} or from wav2vec2 models trained as classifier~\cite{wav2vec_sup_unsup,wav2vec2} have also been explored for accented ASR.    
Many simple ways of fusing accent information with the input speech have been previously investigated. This fusion can either be a sum~\cite{jain2018improved, embed1, wav2vec_sup_unsup}, a weighted sum~\cite{wav2vec2} or a concatenation~\cite{wav2vec_sup_unsup,embed2}. Few works also explore the possibility of merging both accent-aware and accent-agnostic techniques within the same model~\cite{10096722}.
Our work also proposes an accent-aware approach. However, unlike prior work that focuses on pre-fetched accent information, we learn accent information embedded within codebooks during training. Additionally, instead of simply concatenating input speech with accent embeddings, we propose a learned fusion of accent information with speech representations using cross-attention. Prior work by \citet{wav2vec2} demonstrates fine-grained integration of accent information. However, our proposed framework integrates accent information as part of end-to-end training resulting in robust adaptation.

\section{Methodology}
\label{sec:method}

\begin{figure}[t!]
    \centering
	\centerline{\includegraphics[scale=0.345]{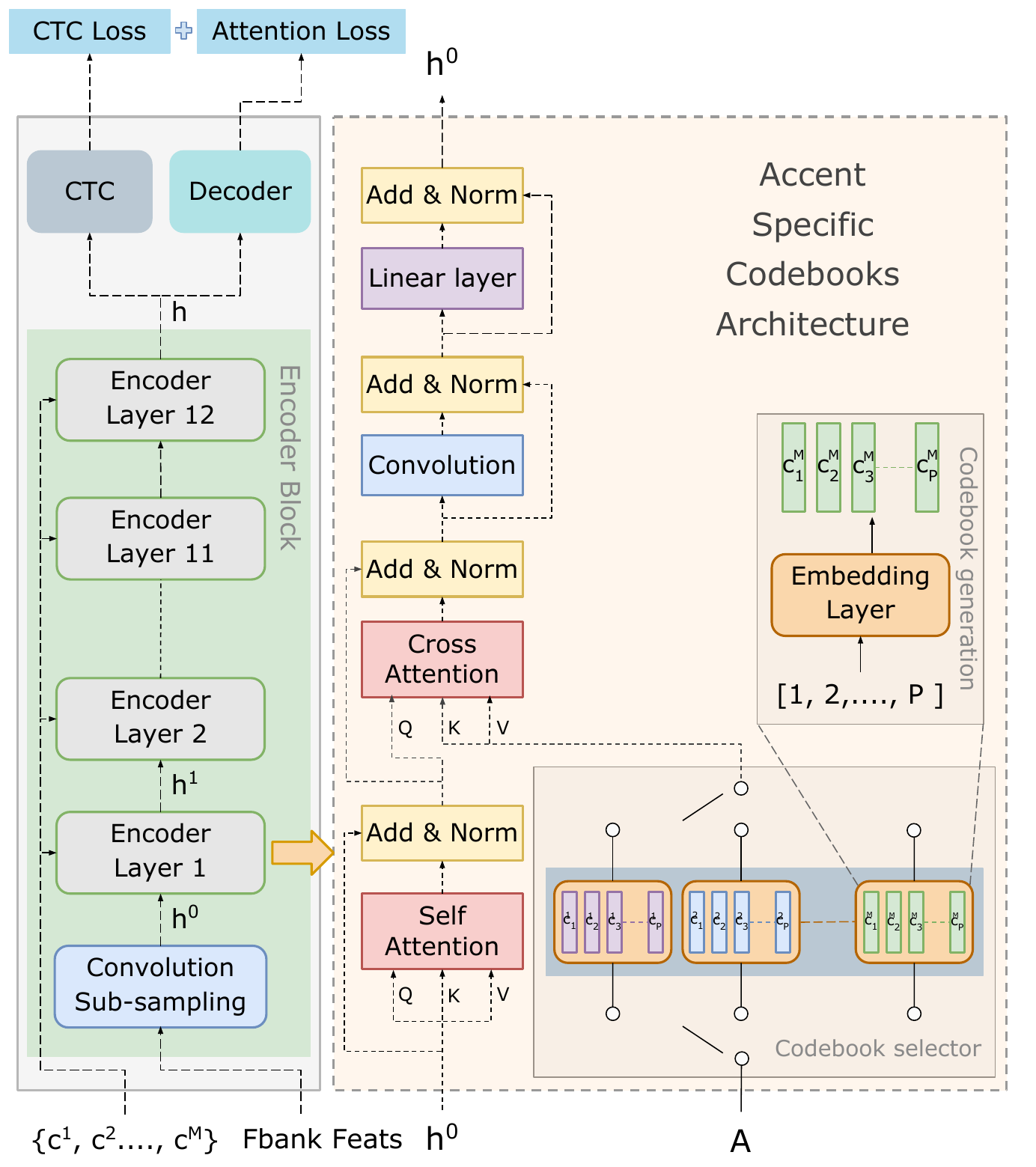}}
	\caption{Overview of our proposed architecture integrating accent codebooks into encoder layers via cross-attention. $A$ represents the accent label for a training instance and $\{ c^1, c^2,\ldots c^M \}$ is the collection of accent-specific codebooks . }
    \vspace{-0.3cm}
	\label{fig:encoder_arch}
\end{figure}
% \begin{figure}[t!]
%     \centering
% 	\centerline{\includegraphics[scale=0.3]{emnlp2023-latex/images/google-drawing-sample.pdf}}
% 	\caption{Overview of our encoder architecture}
% 	\label{fig:encoder_arch}
% \end{figure}

\paragraph{Base model.} Our base architecture uses the standard joint CTC-Attention framework~\cite{jca} with an encoder (\textsc{Enc}), a decoder (\textsc{Dec-Att}), and a Connectionist Temporal Classification (CTC)~\cite{ctc} module (\textsc{Dec-Ctc}). For a given speech input $\mathbf{x}=\{x_1,\ldots,$ $x_T\}$, the encoder $\textsc{Enc}$ generates contextualized  representations  $\mathbf{h}=\textsc{Enc}(\mathbf{x})=\{ h_1,\ldots,h_T \}$. The encoder representations $\mathbf{h}$ are further used by \textsc{Dec-Att} and \textsc{Dec-Ctc} to jointly predict the output token sequence $\mathbf{y}=\{y_1,\ldots,y_j,\ldots,y_U\}$. \textsc{Dec-Att} is an autoregressive decoder that maximizes the conditional likelihood of producing an output token $y_j$ given $\mathbf{h}$ and the previous labels $y_1,\ldots,y_{j-1}$. In contrast, \textsc{Dec-Ctc} uses CTC to maximize the likelihood of $\mathbf{y}$ given $\mathbf{h}$ by marginalizing over all alignments. The encoder is implemented using Conformer layers~\cite{conformer} and the decoder is implemented using Transformer layers~\cite{transformers}.

For our proposed technique, we introduce the following three essential modifications to the base architecture:  
\begin{enumerate*}[label=\roman*)]
    \item Constructing codebooks that can encode accent-specific information (Section~\ref{subsec:codebook}).
    \item Enabling fine-grained integration of accent information with a Conformer-based ASR model using cross-attention (Section~\ref{subsec:asr_enc}).
    \item Modifying beam search decoding for inference in the absence of accent labels at test-time (Section~\ref{subsec:inference}).
\end{enumerate*} 

%Section~\ref{subsec:codebook} provides details about the generation of codebooks and their selection mechanism for a particular example. Figure~\ref{fig:encoder_arch} and section~\ref{subsec:asr_enc} illustrate our proposed encoder architecture that incorporates codebooks using the cross-attention technique. Section~\ref{subsec:inference} outlines our changes to beam search to handle the absence of accent labels during inference. 

\subsection{Codebook Construction}
\label{subsec:codebook}

Consider $M$ seen accents, which are observed during training. We  generate $M$ codebooks, one per accent, where the $i^{\text{th}}$ codebook learns latent codes specific to the $i^{\text{th}}$ accent. During training, we use a deterministic gating scheme to select   the codebook specific to the underlying accent of the training example. To support the selection of a single accent codebook during inference, when the accent labels for the test utterances are unknown, we modify the beam-search decoder to search across all seen accents. We found such a hard gating to be critical to achieve ASR performance improvements. In Section~\ref{sec:analysis}, we compare the proposed model with a soft gating mechanism that works with a standard beam-search decoding.

%the codebook selector functions as a gate, allowing only the codebook representing accent $A$ to be used within the encoder. 
Each codebook contains $P$ $d$-dimensional vectors that we refer to as codebook entries. The entries belonging to the $i^{\text{th}}$ codebook are generated as:
\vspace{-0.1cm}
\[
\mathbf{c}^i = \{c^i_1,\ldots, c^i_{P}\} = \texttt{Embedding}([1,2,\ldots,P])
\]
where \texttt{Embedding} is a standard embedding layer. In the following sections, we use $ \mathbf{c} = \{c_1,\ldots, c_{P}\} $ to refer to the codebook corresponding to the underlying accent label for a given training example. 
%accent $A$, as for the current example, the encoder does not utilize the entries from other codebooks.

\subsection{Encoder with Accent Codebooks}
\label{subsec:asr_enc}

Figure~\ref{fig:encoder_arch} illustrates the overall architecture with the proposed integration of a codebook into each encoder layer via a cross-attention sub-layer. We will refer to this new accent-aware encoder module as $\textsc{Enc}_a = \{\textsc{Enc}^1_a,\ldots,\textsc{Enc}^L_a\}$ that consists of a stack of $L$ identical Conformer layers. The $i^{\text{th}}$ encoder layer $\textsc{Enc}^i_a$ takes both $\mathbf{h}^{i-1}$ and $\mathbf{c}$ as inputs, and produces $\mathbf{h}^i$ as output. Codebook $\mathbf{c}$ is shared across all the encoder layers. All the vectors involved in the computation of attention scores are $d=256$ dimensional. 

A cross-attention sub-layer integrates accent information from codebook $\mathbf{c}$ into each encoder layer. This sub-layer takes both self-attended contextualized representations $\mathbf{H}$ and the codebook $\mathbf{c}$ as its inputs and generates codebook-specific information relevant to the speech frames of this contextual representation. More formally, the operations within encoder layer $\textsc{Enc}^i_a$ can be written as follows:
%Our proposed accented ASR system modifies the encoder with a new cross-attention module that takes as input, codebook $\mathbf{c}=\{c_1, \ldots,c_P\}$, in addition to the input speech $\mathbf{x}$. This results in accent-aware encoder representation $\mathbf{h}$ from this modified encoder, which we call $\textsc{Enc}_a(\mathbf{x},\mathbf{c})$ that acts on both $\mathbf{x}$ and $\mathbf{c}$ as inputs. 

%The encoder module $\textsc{Enc}_a$  consists of a stack of $L$ identical Conformer layers. Let $\textsc{Enc}^i_a$ be the $i^{\text{th}}$ encoder layer that takes $\mathbf{h}^{i-1}$ and $\mathbf{c}$ as inputs, and produces $\mathbf{h}^i$ as output. Codebook $\mathbf{c}$ is shared across all the encoder layers. All the vectors involved in the computation of attention scores are $d=256$ dimensional vectors. The overall architecture of $\textsc{Enc}_a$ is illustrated in Figure~\ref{fig:encoder_arch}. 

%We introduce a cross-attention module amidst the standard blocks of a Conformer layer. This module takes as input the contextualized representation, $\mathbf{H}$ obtained from the self-attention block of the encoder layer and generates codebook-backed information $\mathbf{I}$ relevant to the frames of this contextual representation. This information can be interpreted as the model's attempt to inject accent-specific information into the contextualized speech representation using codebook entries. The new encoder layers in $\textsc{Enc}_{a}$, with cross-attention over accent information, can be written as:
\vspace{-0.3cm}
\begin{align*}
    \mathbf{\hat{H}} &= \mathrm{MultiHeadAttn}_{\text{self}}(\mathbf{h}^{i-1},\mathbf{h}^{i-1},\mathbf{h}^{i-1}) \nonumber \\[-2pt]
    \mathbf{H} &= \mathrm{NormLayer}_{\text{self}}(\mathbf{h}^{i-1} + \mathbf{\hat{H}}) \nonumber \\[-2pt]
    \textcolor{Plum}{\mathbf{\hat{C}}} &= \textcolor{Plum}{\mathrm{MultiHeadAttn}_{\text{cb}}(\mathbf{H},\mathbf{c},\mathbf{c})} \nonumber \\[-2pt]
    \textcolor{Plum}{\mathbf{C}} &= \textcolor{Plum}{\mathrm{NormLayer}_{\text{cb}}(\mathbf{H} + \mathbf{\hat{C}})} \nonumber \\[-2pt]
    \mathbf{\hat{J}} &= \mathrm{Convolution}(\mathbf{C}) \nonumber \\[-2pt]
    \mathbf{J} &= \mathrm{NormLayer}_{\text{conv}}(\mathbf{C} + \mathbf{\hat{J}}) \nonumber \\[-2pt]
    \mathbf{\hat{h}^i} &= \mathrm{Linear}_{\text{pw}}(\mathbf{J}) \nonumber \\[-2pt]
    \mathbf{h}^i &= \mathrm{NormLayer}_{\text{linear}}(\mathbf{J} + \mathbf{\hat{h}}^i) \label{eq:enc_layer}
\end{align*}
\noindent where the equations colored in \textcolor{Plum}{purple} highlight our changes to the standard Conformer encoder layer. The above equations can be viewed as a stack of four independent blocks, each having a residual connection and being separated by layer normalization.

$\mathrm{MultiHeadAttn}(Q,K,V)$ refers to a standard multi-head attention module~\cite{transformers} with $Q$, $K$ and $V$ denoting queries, keys and values respectively. $\mathrm{MultiHeadAttn}_{\text{self}}$ is a self-attention module where each frame of $\mathbf{h}^{i-1}$ attends to every other frame, thus adding contextual information. $\mathrm{Convolution}$ is a stack of three convolution layers: a depth-wise convolution sandwiched between two point-wise convolutions, each having a single stride. The input and output of   the $\mathrm{Convolution}$ block are  $d$-dimensional vectors. A position-wise feed-forward layer $\mathrm{Linear}_{\text{pw}}$ is made up of two linear transformations with a ReLU activation. This takes a $d$-dimensional output from the convolution module as input and produces a $d$-dimensional output vector with a hidden layer of 2048 dimensions.

$\mathrm{MultiHeadAttn}_{\text{cb}}(\mathbf{H}, \mathbf{c}, \mathbf{c})$ is our proposed cross-attention module over codebook entries, where each frame of input $\mathbf{H}$ attends to all the entries in the codebook $\mathbf{c}$ to generate attention scores. These attention scores are further used to generate frame-relevant information $\mathbf{\hat{C}}$ as a weighted average of codebook entries. We elaborate further on the attention computation for a single attention head in $\mathbf{\hat{C}}$.%
\footnote{In our experiments, self-attention uses four attention heads for the encoder and cross-attention with the codebook uses a single attention head.}
Let $\mathbf{H}_{j}$ refer to the $j^{\text{th}}$ frame in $\mathbf{H}$. The attention distribution $\{\alpha_{j,1},\ldots \alpha_{j,P}\}$, where $\alpha_{j,k}$ is the attention probability given by $\mathbf{H}_{j}$ to the $k^{\text{th}}$ codebook entry, is computed as:
\vspace{-0.05cm}
\begin{align*}
    & \{\alpha_{j,1}, \alpha_{j,2},\ldots,\alpha_{j,P}\} = \\ &
    \mathrm{softmax}\left(\frac{(W^i_{q} \mathbf{H}_j)(W^i_{k} \mathbf{c})^T}{\sqrt{d}} \right)( W^i_{v} \cdot \mathbf{c} )
\end{align*}
\vspace{-0.05cm}
\noindent where $W^i_{q}$, $W^i_{k}$ and $W^i_{v} \in \mathbb{R}^{d\times d}$ are learned projection matrices. These attention scores are further used to generate the weighted average of codebook entries in $\mathbf{\hat{C}}$:
\vspace{-0.2cm}
\begin{align*}
    \mathbf{\hat{C}}_{j} = \sum_{k=1}^{P} \alpha_{j,k} \cdot \mathbf{c}_{k}
\end{align*}

%This attention mechanism allows the model to decide what entries from the codebook it wants to add to each frame. 
The cross-attention sublayer is further modified with a residual connection and layer normalization to generate the final codebook-infused representations in $\mathbf{C}$. 
%containing speech content fused with information learned through the codebook. 

\begin{algorithm}[h]
\DontPrintSemicolon
\setstretch{0.975}
\Input{ 
    $\mathbf{x}$: speech input \\
    $\mathcal{V}$: list of vocabulary tokens \\
    $n_{\max}$: maximum hypothesis length \\
    $k$: maximum beam width \\
    $y$: output prediction so far\\
    $\texttt{score}_{A}(.,.,.)$: scoring function\\
}
$ \mathcal{B}_0 $ = \colorbox{green!25}{ $ \{ \langle 0, \texttt{<sos>}, 1 \rangle \ldots , \langle 0, \texttt{<sos>}, M \rangle \} $ } \\ 
\For{ $ t \in \{ 1, \ldots, n_{\max} - 1 \} $ }{
    $ \mathcal{B} \leftarrow \phi $ \\
    \For{ \colorbox{green!25}{ $ \langle s, y, A \rangle $ } $ \in \mathcal{B}_{t-1} $ }{
    
        \If{ $ y.\text{\normalfont last()} == \texttt{<eos>} $ }{
            $ \mathcal{B}\text{\normalfont .add}( \langle s, y, A \rangle ) $ \\
            \text{\normalfont continue} 
        }

        \For{ $ v \in \mathcal{V} $ }{
            $ s \leftarrow $ \colorbox{green!25}{$ \texttt{score}_{A}( \mathbf{x}, y \circ v, A ) $ } \\
            $ \mathcal{B}.\text{\normalfont add}( \langle s, y \circ v, A \rangle ) $
        }
    }
    $\mathcal{B}_t = \mathcal{B}.\text{\normalfont top}(k) $ 
}
\Return{ $\mathcal{B}$ }
\caption{ Inference algorithm that performs joint beam-search over all accents. Our modifications to the standard beam search~\cite{tacl_a_00346} are \colorbox{green!25}{highlighted}. Each beam entry is a triplet $ \langle s, y, A \rangle $ where $A$ refers to a seen accent. $\texttt{score}_{A}()$ is a modified scoring function which uses the codebook for accent $A$ during the forward pass.}
\label{alg:inf}
\end{algorithm} 

\subsection{Modified Beam-Search Algorithm}
\label{subsec:inference}

Since we do not have access to accent labels at test-time, we   rely on either using a classifier to predict the accent or modifying beam-search to accommodate the prediction streams generated by all seen accent choices. Due to a large imbalance in the accent distribution during training with certain seen accents dominating the training set, we find the classifier to be ineffective during inference. We elaborate on this further in Section~\ref{sec:analysis}. 

Figure~\ref{alg:inf} shows our inference algorithm that performs a joint beam search over all the seen accents. Each beam entry is a triplet that expands each hypothesis using each seen accent. Scores for each seen accent are computed using a forward pass through our ASR model by invoking the codebook specific to the accent. The beam width threshold $k$ is then applied to expanded predictions across all seen accents.
%While splitting the beam amongst the seen accents and performing beam search independently on each of them work, it results in beam slots being wasted on accents that are irrelevant. A better choice is to perform joint beam search over all the seen accents at once and let the model's scoring function guide the beam allocation amongst the seen accents. 

\section{Experimental Setup}
\label{sec:setup}

\subsection{Datasets}
\label{subsec:dataset}
All our experiments are conducted on the \mcv dataset extracted from the ``validated" split of the Mozilla Common Voice English (\texttt{en}) corpus~\cite{mcv}. Overall, 14 English accents present in \mcv are divided into two groups of \emph{seen} and \emph{unseen} accents. Table \ref{table:accents} lists the accents belonging to each of these groups.

\begin{table}[h]
    \centering
    \setlength\extrarowheight{-0.5pt}
    \resizebox{\linewidth}{!}{
    \begin{tabular}{| P{0.8cm} P{0.9cm} | P{1.1cm} P{0.8cm} P{1.1cm} P{0.8cm} |  }
    \hline
    \multicolumn{2}{|c|}{\textbf{Seen Accents}} & \multicolumn{4}{c|}{\textbf{Unseen Accents}} \\
    \hline
    \small{Australia} & \scriptsize{(AUS)} & \small{Africa} & \scriptsize{(AFR)} & \small{Malaysia} & \scriptsize{(MAL)} \\
    \small{Canada} & \scriptsize{(CAN)} & \small{New Zealand}  & \scriptsize{(NWZ)} & \small{Hong Kong} &  \scriptsize{(HKG)} \\
    \small{England} & \scriptsize{(GBR)} & \small{Philippines}  & \scriptsize{(PHL)} & \small{India} & \scriptsize{(IND)} \\
    \small{Scotland} & \scriptsize{(SCT)} & \small{Singapore} &  \scriptsize{(SGP)} & \small{Ireland}  & \scriptsize{(IRL)} \\
    \small{US} & \scriptsize{(USA)} & \small{Wales} & \scriptsize{(WLS)} & & \\
    \hline
    \end{tabular}
    }
    \caption{List of $5$ seen and $9$ unseen accents in \mcv corpus.}
    \label{table:accents}
    \vspace{-0.2cm}
\end{table}

%To train our ASR systems, we create a subset \textsc{mcv\_accent-100} from \textsc{mcv\_accent}. 
We create train, dev, and test splits that are speaker-disjoint. We construct two train sets, \textsc{mcv\_accent-100} and \mcv-600, comprising approximately $100$ hours and $620$ hours of labeled accented speech, respectively. Since \mcv consists of many utterances that correspond to the same underlying text prompts, a careful division into dev/test sets that are disjoint in transcripts from the train set was performed. We train and validate the ASR models only on the seen accents, while the test data consists of both seen and unseen accents. The detailed statistics of our datasets are given in Table~\ref{table:asr_data}. For a quick turnaround, most of our experiments were conducted on \textsc{mcv\_accent-100}. For experiments in Section~\ref{sec:bigmcv}, we use the $620$-hour \mcv-600. All the data splits mentioned above are available in our codebase, which should enable direct comparisons across accent adaptation techniques. More details about the construction of the datasets are provided in Appendix~\ref{app:data}.

\begin{table}[h]
    \centering
    \resizebox{\linewidth}{!}{
        \begin{tabular}{ | c | c | c | c | c | }
            \hline
            \multirow{3}{*}{\textbf{Accent}} & \multicolumn{2}{c|}{\multirow{2}{*}{\makecell{ Train \\ (in hours) }}} & \multirow{3}{*}{ \makecell{ Dev  \\ (in hours)}} & \multirow{3}{*}{ \makecell{ Test  \\ (in hours)}} \\
            & \multicolumn{1}{c}{} & & & \\
            \cline{2-3}
            & \small{\textsc{ma-100}} & \small{\textsc{ma-600}} &  &  \\
            \hline
            \small{Australia} & \small{6.95} & \small{45.36} & \small{4.33} & \small{0.46}  \\
            \small{Canada} & \small{6.79}  & \small{41.13} & \small{1.16} & \small{1.21}  \\
            \small{England} & \small{19.51}  & \small{119.9} & \small{3.22} & \small{1.65}  \\
            \small{Scotland} & \small{2.69}  & \small{16.21} & \small{0.23} & \small{0.16}  \\
            \small{US} & \small{64.12}  & \small{400.1} & \small{8.32} & \small{4.87}  \\
            \hline 
            \small{Africa} & \multicolumn{2}{c|}{\small{--}} & \small{--} & \small{1.71}  \\
            \small{Hongkong} & \multicolumn{2}{c|}{\small{--}} & \small{--} & \small{0.52}  \\
            \small{India} & \multicolumn{2}{c|}{\small{--}} & \small{--} & \small{0.58}  \\
            \small{Ireland} & \multicolumn{2}{c|}{\small{--}} & \small{--} & \small{1.94}  \\
            \small{Malaysia} & \multicolumn{2}{c|}{\small{--}} & \small{--} & \small{0.39}  \\
            \small{Newzealand} & \multicolumn{2}{c|}{\small{--}} & \small{--} & \small{2.11}  \\
            \small{Philippines} & \multicolumn{2}{c|}{\small{--}} & \small{--} & \small{0.90}  \\
            \small{Singapore} & \multicolumn{2}{c|}{\small{--}} & \small{--} & \small{0.64}  \\
            \small{Wales} & \multicolumn{2}{c|}{\small{--}} & \small{--} & \small{0.27}  \\
            \hline
            
        \end{tabular}
    }
\caption{Data splits for experiments. \textsc{ma-100} and \textsc{ma-600} refers to datasets \textsc{mcv\_accent-100} and \textsc{mcv\_accent-600}  datasets respectively. }
\label{table:asr_data}
    \vspace{-0.2cm}
\end{table}

\subsection{Models and Implementation Details}
\label{sec:exp_setup}

We use the ESPnet toolkit~\cite{espnet} for all our ASR experiments. %In all these experiments, we use the dataset proposed in section~\ref{subsec:dataset}. 
As is standard practice, we further add 3-way speed perturbation to our dataset before training. We use the default configurations specified in the  \texttt{\small{train\_conformer}}\texttt{\small{.yaml}} file provided in the ESPnet toolkit\footnote{\url{https://tinyurl.com/2wexthds}} to train a Conformer model comprising 12 encoder layers and 6 decoder layers using joint CTC-Attention loss~\cite{jca}. We use four attention heads to attend over 256-dimensional tensors. The position-wise linear layer operates with $2048$ dimensions. We train the model for $50$ epochs using $80$-dimensional filter-bank features with pitch.  
 In all our experiments, we apply a stochastic depth rate of $0.3$ which we found to yield an absolute $2$\% WER improvement, compared to a baseline system without this regularization enabled. During inference, we use a two-layer RNN language model~\cite{rnnlm} trained for 20 epochs with a batch size of 64. We conducted all our experiments on NVIDIA RTX A6000 GPUs.

\section{Experiments and Results}
\label{sec:res}

\begin{table*}[ht!]
\centering
\resizebox{\linewidth}{!}{
    \begin{tabular}{ l|c|cc|ccccc|ccccccccccc }
        \hline
        \hline
        
        \multicolumn{1}{c|}{\multirow{2}{*}{\textbf{\small{Method}}}} & \multicolumn{3}{c|}{\textbf{\footnotesize{Aggregated}}} & \multicolumn{5}{c|}{\textbf{\footnotesize{Seen Accents}}} & \multicolumn{9}{c}{\textbf{\footnotesize{Unseen Accents}}} \\
        
        \cline{2-18}

        & \textbf{\scriptsize{All}} & \textbf{\scriptsize{Seen}} & \textbf{\scriptsize{Unseen}} & \textbf{\scriptsize{AUS}} & \textbf{\scriptsize{CAN}} & \textbf{\scriptsize{UK}} &
        \textbf{\scriptsize{SCT}}  & \textbf{\scriptsize{US}}  & 
        
        \textbf{\scriptsize{AFR}} & \textbf{\scriptsize{HKG}} & \textbf{\scriptsize{IND}} & \textbf{\scriptsize{IRL}} & \textbf{\scriptsize{MAL}} & \textbf{\scriptsize{NWZ}} & \textbf{\scriptsize{PHL}} & \textbf{\scriptsize{SGP}} & \textbf{\scriptsize{WLS}}  \\
        
        \hline
        
        \small{Trans. \cite{speech-transformer}} & \footnotesize{22.7} &  \footnotesize{17.3} &  \footnotesize{28.0} &  \footnotesize{18.1} & \footnotesize{17.8} & \footnotesize{19.7} & \footnotesize{18.5} & \footnotesize{16.3} & \footnotesize{25.9} & \footnotesize{32.0} & \footnotesize{35.4} & \footnotesize{25.3} & \footnotesize{36.2} & \footnotesize{23.8} & \footnotesize{31.5} & \footnotesize{38.8} & \footnotesize{21.0} \\
    
        \small{Conf. \cite{conformer}} & \footnotesize{18.9} &  \footnotesize{14.0} &  \footnotesize{23.7} &  \footnotesize{13.8} & \footnotesize{\textbf{15.0}} & \footnotesize{15.7} & \footnotesize{\textbf{13.4}} & \footnotesize{13.3} & \footnotesize{21.5} & \footnotesize{27.2} & \footnotesize{29.4} & \footnotesize{21.4} & \footnotesize{32.2} & \footnotesize{19.9} & \footnotesize{26.1} & \footnotesize{34.7} & \footnotesize{17.9} \\
    
        \small{I-vector   \cite{ivec}} & \footnotesize{18.9} &  \footnotesize{14.1} & \footnotesize{23.6} &  \footnotesize{13.9} & \footnotesize{15.0} & \footnotesize{16.1} & \footnotesize{14.6} & \footnotesize{13.3} & \footnotesize{21.7} & \footnotesize{27.2} & \footnotesize{29.5} & \footnotesize{21.2} & \footnotesize{\textbf{31.7}} & \footnotesize{19.3} & \footnotesize{27.2} & \footnotesize{\textbf{33.8}} & \footnotesize{18.0} \\
    
        \small{MTL \cite{asr_clf}} &\footnotesize{18.9} &  \footnotesize{14.1} & \footnotesize{23.7} & \footnotesize{14.7} & \footnotesize{15.1} & \footnotesize{16.1} & \footnotesize{13.7} & \footnotesize{13.2} & \footnotesize{21.8} & \footnotesize{28.1} & \footnotesize{\textbf{29.1}} & \footnotesize{21.5} & \footnotesize{33.0} & \footnotesize{19.4} & \footnotesize{26.5} & \footnotesize{34.2} & \footnotesize{18.1}  \\
    
        \small{DAT} \cite{bobw} & \footnotesize{\textbf{18.7}} &  \footnotesize{\textbf{14.0}} & \footnotesize{\textbf{23.4}} &  \footnotesize{\textbf{13.3}} & \footnotesize{15.3} & \footnotesize{\textbf{15.7}} & \footnotesize{15.5} & \footnotesize{\textbf{13.1}} & \footnotesize{\textbf{21.1}} & \footnotesize{\textbf{27.0}} & \footnotesize{29.5} & \footnotesize{\textbf{21.1}} & \footnotesize{32.2} & \footnotesize{\textbf{19.2}} & \footnotesize{\textbf{26.0}} & \footnotesize{34.4} & \footnotesize{\textbf{17.9}} \\
    
        \hline
    
        \small{\textsc{CA}  } & \cellcolor{green!20} \footnotesize{\textbf{18.2}$\dagger$} &  \cellcolor{green!20} \footnotesize{\textbf{13.6}} & \cellcolor{green!20} \footnotesize{\textbf{22.9}} &  \cellcolor{green!20} \footnotesize{\textbf{11.5}} & \cellcolor{green!20} \footnotesize{\textbf{14.8}} & \cellcolor{green!20} \footnotesize{\textbf{14.9}} & \cellcolor{green!20} \footnotesize{\textbf{9.7}} & \cellcolor{green!20} \footnotesize{\textbf{13.1}} & \cellcolor{green!20} \footnotesize{\textbf{21.0}} & \cellcolor{green!20} \footnotesize{\textbf{25.7}} & \cellcolor{green!20} \footnotesize{\textbf{29.1}} & \cellcolor{green!20} \footnotesize{\textbf{20.7}} & \cellcolor{green!20} \footnotesize{\textbf{30.9}} & \cellcolor{green!20} \footnotesize{\textbf{18.5}} & \cellcolor{green!20} \footnotesize{\textbf{25.8}} & \cellcolor{green!20} \footnotesize{\textbf{33.7}} & \cellcolor{green!20} \footnotesize{\textbf{17.9}} \\
        
        \hline
        \hline
    \end{tabular}
}
\caption{Comparison of the performance (WER \% ) of our architecture  (codebook attend (CA)) with baseline and other techniques on the MCV-ACCENT-100 dataset. Numbers in bold denote the best across baselines, and the green highlighting \colorbox{green!20}{$\phantom{x}$} denotes the best WER across all experiments. Ties are broken using overall WER. $\textsc{CA}$: Codebook attend - cross-attention applied at all layers with $50$ entries in each learnable  codebook. $\dagger$ indicates statistically significant results compared to DAT (at $p$ <0.001 using MAPSSWE  test \cite{mapsswe}).}
\label{table:res_asr}
    \vspace{-0.2cm}

\end{table*}

Table~\ref{table:res_asr} shows word error rates (WERs) comparing our best system (codebook attend (CA) system) with five approaches:
\begin{enumerate*}
    \item Transformer baseline~\cite{speech-transformer}.
    \item Conformer baseline~\cite{conformer}.
    \item Adding i-vector features \cite{ivec}\footnote{Every frame in i-vectors represents a second. Since the input filterbank features span 25msec, we repeat the same i-vector frame for four consecutive feature frames. Adding, as opposed to concatenating, the i-vector frames was found to perform better.} to the filterbank features, as input to the Conformer baseline.
    \item Conformer jointly trained with an accent classifier using multi-task learning~\cite{asr_clf}.
    \item Conformer with Domain Adversarial Training (DAT)~\cite{bobw}, with an accent classifier at the $10^{\text{th}}$ encoder layer.
\end{enumerate*}

From Table~\ref{table:res_asr}, we observe that the Conformer baseline performs significantly better than the Transformer baseline. Adding i-vectors and multi-task training with an auxiliary accent classifier objective perform equally well and are comparable to the Conformer baseline, while using DAT improves over the Conformer baseline. Our system significantly outperforms DAT (at $p<0.001$ using the MAPSSWE test~\cite{mapsswe}) and achieves the lowest WERs across all the seen and unseen accents. We use $50$ codebook entries for each accent and incorporate accent codebooks into each of the $12$ encoder layers. Unless specified otherwise, we will use this configuration   in all subsequent experiments. Further ablations of these choices will be detailed in Section~\ref{subsec:ablation}.
%While P1 and P2 perform comparable to the best Baseline B2, P3 performs better than both P1 and P2. On overall WER, our cross-attention technique (\textbf{T1}) achieves an overall 3\% relative reduction in WER compared to baselines and other existing approaches.

%In the sections that follow(i.e sections~\ref{subsec:zero-shot},~\ref{subsec:diff_inf} and \ref{subsec:single_inf}), we use the model trained using T1's architecture to perform our analysis.

\subsection{Zero-shot Transfer}
\label{subsec:zero-shot}

\begin{table}[htb]
\centering
\resizebox{\linewidth}{!}{
\begin{tabular}{ | l | c | cccccc | }
    \hline
    \multicolumn{1}{|c|}{\multirow{2}{*}{\textbf{\small{Method}}}} & \multirow{2}{*}{\textbf{\small{All}}} & \multicolumn{6}{c|}{\textbf{\footnotesize{Accents}}} \\
    
    \cline{3-8}
    
    &  &  \textbf{\scriptsize{ARA}} & \textbf{\scriptsize{HIN}} & \textbf{\scriptsize{KOR}} &
    \textbf{\scriptsize{MAN}}  & \textbf{\scriptsize{SPA}}  & 
    \textbf{\scriptsize{VIA}}  \\

    \hline

   \small{Conformer} & \footnotesize{33.3} & \footnotesize{30.4} & \footnotesize{30.4} & \footnotesize{26.9} & \footnotesize{37.9} & \footnotesize{30.3} & \footnotesize{43.5}  \\

   \small{I-vector} & \footnotesize{33.6} &  \footnotesize{31.0} & \footnotesize{31.2} & \footnotesize{27.2} & \footnotesize{38.0} & \footnotesize{30.4} & \footnotesize{43.9} \\

  \small{MTL} & \footnotesize{33.4} &  \footnotesize{30.4} & \footnotesize{30.6} & \footnotesize{26.9} & \footnotesize{38.7} & \footnotesize{30.1} & \footnotesize{43.7} \\

   \small{DAT} & \footnotesize{33.5} &  \footnotesize{30.7} & \footnotesize{30.8} & \footnotesize{26.8} & \footnotesize{38.3} & \footnotesize{30.1} & \footnotesize{43.9} \\

   \hline

   \small{\textsc{CA}  } & \footnotesize{\textbf{32.6}$\dagger$} &  \footnotesize{\textbf{29.5}} & \footnotesize{\textbf{30.4}} & \footnotesize{\textbf{26.2}} & \footnotesize{\textbf{37.1}} & \footnotesize{\textbf{29.3}} & \footnotesize{\textbf{42.8}} \\
    
    \hline
\end{tabular}
}
\caption{Comparison of the zero-shot performance (WER \%) of our architecture with other techniques on L2Arctic dataset. $\dagger$ indicates a statistically significant improvement ($p$ <0.001 using MAPSSWE test) using codebook attend (\textsc{CA}) w.r.t. the Conformer baseline. }
\label{table:res_l2arctic}
    \vspace{-0.2cm}

\end{table}

To further validate the efficacy of our proposed approach using accent-specific codebooks, we perform zero-shot evaluations on the L2Arctic dataset. We note here that we do not use any L2Arctic data for finetuning; our ASR model is trained on \mcv-100. Such a zero-shot evaluation helps ascertain whether our codebooks transfer well across datasets.
%So far, we have been comparing the experiments using an in-domain test set derived from Commonvoice. To further show the effectiveness of our approach on out-of-domain examples, we perform zero-shot inference on the L2Arctic dataset. 
The L2Arctic dataset \cite{l2arctic} comprises English utterances spanning six non-native English accents namely Arabic (\texttt{ARA}), Hindi (\texttt{HIN}), Korean (\texttt{KOR}), Mandarin (\texttt{MAN}), Spanish (\texttt{SPA}), and Vietnamese (\texttt{VIA}). Table~\ref{table:res_l2arctic} shows WERs achieved by our system in comparison to the baseline and other techniques. Our proposed method significantly outperforms all these approaches on every single accent ($p<0.001$ using the MAPSSWE test \cite{mapsswe}). 

\subsection{Effect of Training Data Size}
\label{sec:bigmcv}

\begin{table}[htb]
\centering
\resizebox{\linewidth}{!}{
\begin{tabular}{ | l | c | c | c | }
    \hline

    \multicolumn{1}{|c|}{\textbf{\small{Method}}} & \textbf{\small{Overall}} & \textbf{\small{Seen}} & \textbf{\small{Unseen}} \\

    \hline

   \small{Conf.~\cite{conformer}} & \footnotesize{9.75} &  \footnotesize{\textbf{6.04}} & \footnotesize{13.46} \\

   \small{I-vector~\cite{ivec}} & \footnotesize{10.05} &  \footnotesize{6.40} & \footnotesize{13.69} \\

   \small{MTL~\cite{asr_clf}} & \footnotesize{10.02} &  \footnotesize{6.33} & \footnotesize{13.70} \\

  \small{DAT~\cite{bobw}} & \footnotesize{9.73} &  \footnotesize{6.12} & \footnotesize{13.33} \\

    \hline
   
    \small{$\textsc{CA}_{L\in (1,\ldots,12)}(P=50)$} & \footnotesize{9.63} &  \footnotesize{6.22} & \footnotesize{13.03} \\

    \small{$\textsc{CA}_{L\in (1,\ldots,12)}(P=200)$} & \footnotesize{9.59} &  \footnotesize{6.20} & \footnotesize{12.98} \\

     \small{$\textsc{CA}_{L\in (1,\ldots,12)}(P=500)$} & \footnotesize{\textbf{9.55}} &  \footnotesize{6.19} & \footnotesize{\textbf{12.92}} \\

   \hline
    
\end{tabular}
}
\caption{Comparison of the performance (WER \%) of our approach with other methodologies on \mcv-600 dataset.
% ~$\textbf{X}$ denotes the best amongst experiments belonging to a particular tabular subsection, and \colorbox{green!20}{$\textbf{X}$} denotes the best across all the experiments. Ties are broken using overall WER values. 
% $\texttt{CA}_{L \in (i,\ldots,j)}(P=k)$: Cross Attention applied at all layers from $i$ to $j$ with $k$ entries per accent codebook. $\texttt{CA}_{L \in (i,\ldots,j)}(P_{frozen}=k)$: Similar to the previous setup, but with codebooks frozen during training
%Accent-wise WER   is shown in Appendix~\ref{app:results}.
 }
\label{table:res_bigger_dataset}
\vspace{-0.2cm}

\end{table}

Table~\ref{table:res_bigger_dataset} compares our proposed system with DAT and Conformer on the $600$-hour \mcv dataset. Compared to the Conformer baseline and the DAT, the proposed CA approach shows a steady improvement over unseen accents, while resulting in a minor drop in performance on the seen accents. %Meanwhile, Our architecture shows a 3\% relative WER reduction on the unseen accents over the baseline.

\subsection{Effect of Number of Parameters}
\label{sec:bigmcv}

\begin{table}[htb]
\centering
\resizebox{\linewidth}{!}{
\begin{tabular}{ | l | c | c | c | c | }
    \hline

    \multicolumn{1}{|c|}{\textbf{\small{Method}}} & \textbf{\small{\# of params}} & \textbf{\small{Overall}} & \textbf{\small{Seen}} & \textbf{\small{Unseen}} \\

    \hline

   \small{Conf.} & \footnotesize{43M} & \footnotesize{18.87} &  \footnotesize{14.05} & \footnotesize{23.67} \\

   \small{Conf. w/ $\uparrow$ encoder units} & \footnotesize{46M} & \footnotesize{18.89} &  \footnotesize{14.02} & \footnotesize{23.74} \\

   \small{Conf. w/ $\uparrow$ attention dim } & \footnotesize{46M} & \footnotesize{18.77} &  \footnotesize{14.02} & \footnotesize{23.51} \\

    \hline
   
   \small{$\textsc{CA}_{L\in (1,\ldots,12)}(P=50)$} & \footnotesize{46M} & \footnotesize{\textbf{18.22}} &  \footnotesize{\textbf{13.57}} & \footnotesize{\textbf{22.86}} \\

   \hline
    
\end{tabular}
}
\caption{Comparison of the performance (WER \%) of our approach with parameter-equivalent variants of the Conformer baseline on \mcv-100.}
\label{table:res_model_params}
\vspace{-0.2cm}
\end{table}

To discount the possibility that improvements using our proposed model could be attributed to an increase in the number of parameters, in Table~\ref{table:res_model_params}, we compare our proposed system with multiple variants of the baseline Conformer model~(referred to as Conf. in Table~\ref{table:res_asr}) where parameters are increased to be commensurate with our proposed model by either
\begin{enumerate*}[label={(\arabic*)}]
    \item Increasing the number of encoder units (from $2048 \rightarrow 2320$) or 
    \item Increasing the dimension used for attention computation (from $256 \rightarrow 272$).
\end{enumerate*}
We observe a slight improvement over the standard baseline when the attention dimension is increased. However, compared to all these baselines, our proposed model still shows a statistically significant improvement at $p$ <0.001.

\subsection{Balanced versus Imbalanced Dataset}
\label{sec:balancedmcv}

\begin{table}[htb]
\centering
\resizebox{\linewidth}{!}{
\begin{tabular}{ | l | c | c | c | }
    \hline

    \multicolumn{1}{|c|}{\textbf{\small{Method}}} & \textbf{\small{Overall}} & \textbf{\small{Seen}} & \textbf{\small{Unseen}} \\

    \hline

   \small{Conformer} & \footnotesize{19.30} &  \footnotesize{14.73} & \footnotesize{23.86} \\

    % \hline
   
   \small{$\textsc{CA}_{L\in (1,\ldots,12)}(P=50)$} & \footnotesize{\textbf{18.88}} &  \footnotesize{\textbf{14.61}} & \footnotesize{\textbf{23.13}} \\

   \hline
    
\end{tabular}
}
\caption{Comparison of the performance (WER \%) of our approach with Conformer baseline on an accent balanced \mcv-100 dataset.
 }
\label{table:res_balanced_dataset}
\vspace{-0.2cm}
\end{table}

To check the effectiveness of our approach on a balanced dataset, in
 Table~\ref{table:res_balanced_dataset}, we compare our proposed system with the Conformer baseline on a 100-hour accent-balanced data split. Even on such a  balanced dataset, our architecture shows a statistically significant improvement (at $p$=0.005) compared to the baseline.
\begin{table}[t!]
\centering
\resizebox{\linewidth}{!}{
\begin{tabular}{ | l | c | c | c | }
    \hline

    \multicolumn{1}{|c|}{\textbf{\small{Method}}} & \textbf{\small{Overall}} & \textbf{\small{Seen}} & \textbf{\small{Unseen}} \\

    \hline

   \small{$\textsc{CA}_{L\in (1,\ldots,12)}(P=25)$} & \footnotesize{18.33} &  \footnotesize{13.76} & \footnotesize{22.89} \\

   \small{$\textsc{CA}_{L \in (1,\ldots,12)}(P=50)$} & 
   \cellcolor{green!20} \footnotesize{\textbf{18.22}} &  \cellcolor{green!20} \footnotesize{\textbf{13.57}} & \footnotesize{\textbf{22.86}} \\

   \small{$\textsc{CA}_{L\in (1,\ldots,12)}(P=100)$} & \footnotesize{18.36} &  \footnotesize{13.85} & \footnotesize{22.86} \\

   \small{$\textsc{CA}_{L\in (1,\ldots,12)}(P=200)$} & \footnotesize{18.41} &  \footnotesize{13.69} & \footnotesize{23.12} \\

   \small{$\textsc{CA}_{L\in (1,\ldots,12)}(P=500)$} & \footnotesize{18.39} &  \footnotesize{13.68} & \footnotesize{23.09} \\

    \hline

   \small{$\textsc{CA}_{L\in (1,\ldots,4)}(P=50)$} & \footnotesize{\textbf{18.30}} &  \footnotesize{13.95} & \cellcolor{green!20} \footnotesize{\textbf{22.64}} \\

   \small{$\textsc{CA}_{L\in (1,\ldots,8)}(P=50)$} & \footnotesize{18.31} &  \footnotesize{\textbf{13.86}} & \footnotesize{22.75} \\
   
   \small{$\textsc{CA}_{L\in (9,\ldots,12)}(P=50)$} & \footnotesize{18.92} &  \footnotesize{14.24} & \footnotesize{23.59} \\
   
   \small{$\textsc{CA}_{L\in (5,\ldots,12)}(P=50)$} & \footnotesize{18.45} &  \footnotesize{13.84} & \footnotesize{23.05} \\
    
    \hline

   \small{$\textsc{CA}_{L\in (1,\ldots,12)}(P_\text{rand}=50)$} & \footnotesize{18.30} &  \footnotesize{13.65} & \footnotesize{22.95} \\

   \hline
    
\end{tabular}
}
\caption{Comparison of the performance (WER \%) of different variants of our architecture.
% ~$\textbf{X}$ denotes the best amongst experiments belonging to a particular tabular subsection, and \colorbox{green!20}{$\textbf{X}$} denotes the best across all the experiments. Ties are broken using overall WER values. 
$\textsc{CA}_{L \in (i,\ldots,j)}(P=k)$: Codebook attention applied at all layers from $i$ to $j$ with $k$ entries per accent codebook. $\textsc{CA}_{L \in (i,\ldots,j)}(P_\text{rand}=k)$: Similar to the previous setup, but with codebooks frozen during training. Accent-wise WER is shown in Appendix~\ref{app:results} and a few select examples are highlighted in Appendix~\ref{app:examples}.}
\label{table:res_ablation}
\vspace{-0.2cm}
\end{table}
%
% This table belongs to section 5.5, but added here to keep it at the top of the page
\begin{table*}[t!]
\centering
\resizebox{\linewidth}{!}{
    \begin{tabular}{ l | ccccc | ccccccccccc  }
        \hline
        
        \multicolumn{1}{c|}{\multirow{2}{*}{\centering \textbf{\small{Accent used}}}} & \multicolumn{5}{c|}{\textbf{\footnotesize{Seen Accents}}} & \multicolumn{9}{c}{\textbf{\footnotesize{Unseen Accents}}} \\
        
        \cline{2-15}
        
         & \textbf{\scriptsize{AUS}} & \textbf{\scriptsize{CAN}} & \textbf{\scriptsize{UK}} &
        \textbf{\scriptsize{SCT}}  & \textbf{\scriptsize{US}}  & 
        \textbf{\scriptsize{AFR}} & \textbf{\scriptsize{HKG}} & \textbf{\scriptsize{IND}} & \textbf{\scriptsize{IRL}} & \textbf{\scriptsize{MAL}} & \textbf{\scriptsize{NWZ}} & \textbf{\scriptsize{PHL}} & \textbf{\scriptsize{SGP}} & \textbf{\scriptsize{WLS}}  \\
        
        \hline
        
         \small{ \texttt{Australia} } & \cellcolor{green!20}  \footnotesize{\textbf{11.5}} & \footnotesize{19.5} & \footnotesize{17.0} & \footnotesize{18.1} & \footnotesize{17.4} & \footnotesize{22.0} & \footnotesize{29.8} & \footnotesize{32.5} & \footnotesize{24.3} & \footnotesize{33.7} & \cellcolor{green!20} \footnotesize{\textbf{18.7}} & \footnotesize{30.1} & \footnotesize{37.8} & \footnotesize{21.1} \\
    
        \small{ \texttt{Canada} } & \footnotesize{20.5} & \cellcolor{green!20} \footnotesize{\textbf{14.7}} & \footnotesize{20.0} & \footnotesize{15.7} & \footnotesize{13.5} & \footnotesize{24.5} & \footnotesize{27.4} & \footnotesize{29.6} & \cellcolor{green!20} \footnotesize{\textbf{21.4}} & \footnotesize{32.7} & \footnotesize{25.7} & \footnotesize{26.6} & \footnotesize{35.4} & \footnotesize{21.8}\\

        \small{ \texttt{England} } &  \footnotesize{13.8} & \footnotesize{17.7} & \cellcolor{green!20} \footnotesize{\textbf{15.0}} & \footnotesize{14.4} & \footnotesize{16.2} & \cellcolor{green!20} \footnotesize{\textbf{21.5}} & \footnotesize{27.0} & \footnotesize{29.9} & \footnotesize{22.0} & \cellcolor{green!20} \footnotesize{\textbf{32.3}} & \footnotesize{21.1} & \footnotesize{27.1} & \footnotesize{34.8} & \cellcolor{green!20} \footnotesize{\textbf{18.0}} \\

        \small{ \texttt{Scotland} } &  \footnotesize{20.7} & \footnotesize{17.8} & \footnotesize{19.1} & \cellcolor{green!20} \footnotesize{\textbf{10.2}} & \footnotesize{16.4} & \footnotesize{24.4} & \footnotesize{28.2} & \footnotesize{33.5} & \footnotesize{22.6} & \footnotesize{34.4} & \footnotesize{25.7} & \footnotesize{29.0} & \footnotesize{36.6} & \footnotesize{21.3} \\

        \small{ \texttt{US} } &  \footnotesize{20.2} & \cellcolor{green!20} \footnotesize{\textbf{14.7}} & \footnotesize{19.4} & \footnotesize{15.5} & \cellcolor{green!20} \footnotesize{\textbf{13.2}} & \footnotesize{23.4} & \cellcolor{green!20} \footnotesize{\textbf{27.0}} & \cellcolor{green!20} \footnotesize{\textbf{28.1}} & \footnotesize{21.7} & \footnotesize{32.4} & \footnotesize{24.7} & \cellcolor{green!20} \footnotesize{\textbf{25.8}} & \cellcolor{green!20} \footnotesize{\textbf{34.3}} & \footnotesize{22.2} \\
        
        \hline
    \end{tabular}
}
\caption{Comparison of the performances (WER\%) of inferences done using fixed accent labels.}
\label{table:res_oracle_single_inf}
\vspace{-0.2cm}
\end{table*}
\subsection{Ablation Studies}
\label{subsec:ablation}

We present two ablation analyses examining the effect of changing the number of accent-specific codebook entries ($P$) and the effect of applying cross-attention at different encoder layers.

The first five rows in Table~\ref{table:res_ablation} refer to the addition of codebooks to all encoder layers via cross-attention with varying accent-specific codebook sizes ($P$) ranging from $25$ to $500$. As $P$ increases, the experiments show improved performance on seen accents but degrades on the unseen accents, indicating that the codebooks begin to overfit to the seen accents. Our best-performing system with $P=50$ performs well on seen accents while also generalizing to the unseen accents. As expected, using  lower-capacity codebooks ($P=25$) shows performance degradation. 
%gives our best-performing system; using a smaller number of codebooks is  As we approach smaller values of $P$, the bottleneck effect results in experiments showing improved performance on both seen and unseen but only up to a particular threshold, after which the performance degrades. 

The next five rows in Table~\ref{table:res_ablation} refer to codebooks with cross-attention introduced at varying encoder layers. The number of codebook entries is fixed at $50$. Since accent effects can be largely attributed to acoustic differences, we see that the early encoder layers closer to the speech inputs benefit most from the codebooks. Adding codebooks only to the last four or eight encoder layers is not   beneficial. 
%The early layers of the encoder are trained to handle raw speech, and hence codebooks have a larger impact when used at lower layers. This is clear when we compare experiments with cross-attention added to the last four layers versus the last eight layers( last two entries in Table~\ref{table:res_ablation}), wherein just by adding cross-attention to a few more penultimate layers, we see a 3\% relative WER improvement. 

Randomly initialized codebooks were observed to be as useful as learnable codebooks for self-supervised representation learning in \citet{chiu2022selfsupervised}. Motivated by this result, we experiment with randomly-initialized accent-specific codebooks that are not learned during training. The last row of Table~\ref{table:res_ablation} shows that random codebooks only cause a slight degradation in performance compared to the best performing system, echoing the observations in~\citet{chiu2022selfsupervised}.
%Inspired by the work of \cite{chiu2022selfsupervised}, we retrain our best architecture with the codebooks frozen during training. The result is shown in the last row of Table~\ref{table:res_ablation}. As was highlighted by Chiu et al. in their work, the frozen codebooks seem to be performing comparably to its counterpart, with the learned codebooks having a slight edge.  

\subsection{Inference with a Single Accent}
\label{subsec:single_inf}

To understand the effectiveness of accent-specific codebooks, we conduct five experiments by committing to a single seen accent during inference. That is, we decode all the test utterances using a fixed accent label. Table~\ref{table:res_oracle_single_inf} shows results from inferring with a single accent across both seen and unseen accents. For the seen accents, the diagonal contains the lowest WERs indicating that the information learned in our codebooks   benefits the accented samples. Furthermore, similar accents, from geographically-close regions, benefit  each other. The New Zealand accented English speech  achieves the best WERs using Australian accent specific codebooks, Hong Kong, Indian, Philippines and Singapore accented test utterances prefer US accented codebooks, and Wales accent achieves its best results using England-specific codebooks.

The WER results achieved by our best-performing system in Table~\ref{table:res_asr} are much lower than the best WER results achieved in these single-accent experiments. This indicates that one cannot directly map an unseen accent to an appropriate seen accent and therefore, making this decision independently for each utterance (as we propose to do in the joint beam search) is crucial. 

% The primary reason for such a difference could be because of labelling of speakers based on location rather than their accent. 

\subsection{Beam-Search Decoding Variants}
\label{subsec:diff_inf}

All the results reported thus far use a joint beam search decoding. Table~\ref{table:res_inference} shows a comparison of our proposed joint beam search (elaborated in Section~\ref{subsec:inference}) with other beam-search variants incurring varying inference overheads. 
\begin{table}[htb]
\centering
\resizebox{\linewidth}{!}{
\begin{tabular}{ | l | c | c | c | c |}
    \hline

    \multicolumn{1}{|c|}{\textbf{\small{Method}}} & \textbf{\small{All}} & \textbf{\small{Seen}} & \textbf{\small{Unseen}} & \textbf{\small{Inference Time}} \\

    \hline

   \small{$\mathbb{B}_0$: Standard beam search} & \small{18.87} & \small{14.05} & \small{23.67} & 1.0 \\

   \hline
   
   \small{$\mathbb{B}_1$: $M$ full beam searches} & \small{\textbf{18.10}} & \small{\textbf{13.48}} & \small{\textbf{22.71}} & $5.02$ \\

   \small{$\mathbb{B}_2$: $M$ split beam searches } & 
   \footnotesize{18.30} & \footnotesize{13.61} & \footnotesize{22.97} & $1.14$ \\

   \small{$\mathbb{B}_3$: Joint beam search } & \footnotesize{18.22} &  \footnotesize{13.57} & \footnotesize{22.86} & $1.16$ \\
   \hline
\end{tabular}
}
\caption{WER (\%) of various inference algorithms described in section~\ref{subsec:diff_inf} on \mcv-100 setup. Inference time gives a relative comparison of the time taken by each decoding variant with the standard beam search as the reference.}
\label{table:res_inference}
  \vspace{-0.2cm}
\end{table}
$\mathbb{B}_0$ in Table~\ref{table:res_inference} refers to a standard beam-search decoding over the Conformer baseline with a beam width of $k$. The setting $\mathbb{B}_1$ and $\mathbb{B}_2$ refer to running beam-search $M$ times, once for each seen accent and picking the best-scoring hypothesis among all predictions. For  $\mathbb{B}_1$ setting, we use a beam width of $k$ for each seen accent. Naturally, this incurs a large decoding overhead with a factor of $M$ increase in inference time and changes the effective beam width to $Mk$. In the   $\mathbb{B}_2$ setting, we divide the beam width into $M$ parts, each occupied by a specific accent, thus making the effective beam width $k/M$. The setting $\mathbb{B}_1$ performs the best, but significantly increases inference overhead. The $\mathbb{B}_2$ setting is efficient but under-performs due to all accents being given an equal number of beam slots leading to eventual under-utilization of beam slots. Our proposed joint inference in the $\mathbb{B}_3$ setting is an effective compromise of $\mathbb{B}_1$ and $\mathbb{B}_2$,  incurring similar inference overheads as $\mathbb{B}_2$ and achieving a performance closer to the $\mathbb{B}_1$ setting. 

%The $i^{\text{th}}$ beam search here refers to using label $i$ as the pseudo accent label for the codebook selection. The hypothesis with the highest score across all the beam searches is returned as the final prediction. I1 and I2 follow this same format but differ in the size of the beam they operate with.I1 runs all its beam searches with a beam size of $k$. I2, on the other hand, divides the entire beam into $M$ equal parts, each to be occupied by a specific accent. I2 could be imagined as running I1 but with a beam size of $k/M$. Although I1 performs best among all three, it comes with massive overhead and is around $M$ times slower than its counterpart. I2 on the other hand has minimal overhead, but suffers from wastage of beam slots, as not all accents perform equally well on a particular utterance.

%To address all these concerns, we infer using \textbf{I3} that runs one joint beam search over all the accents. This could be visualized as a game of survival of the fittest, where the accents that perform terribly are eliminated in the early beam steps. More details about this algorithm can be found in section~\ref{subsec:inference}.  

\section{Discussion and Analysis}
\label{sec:analysis}

\begin{figure}[t!]
    \centering
\centerline{\includegraphics[scale=0.47]{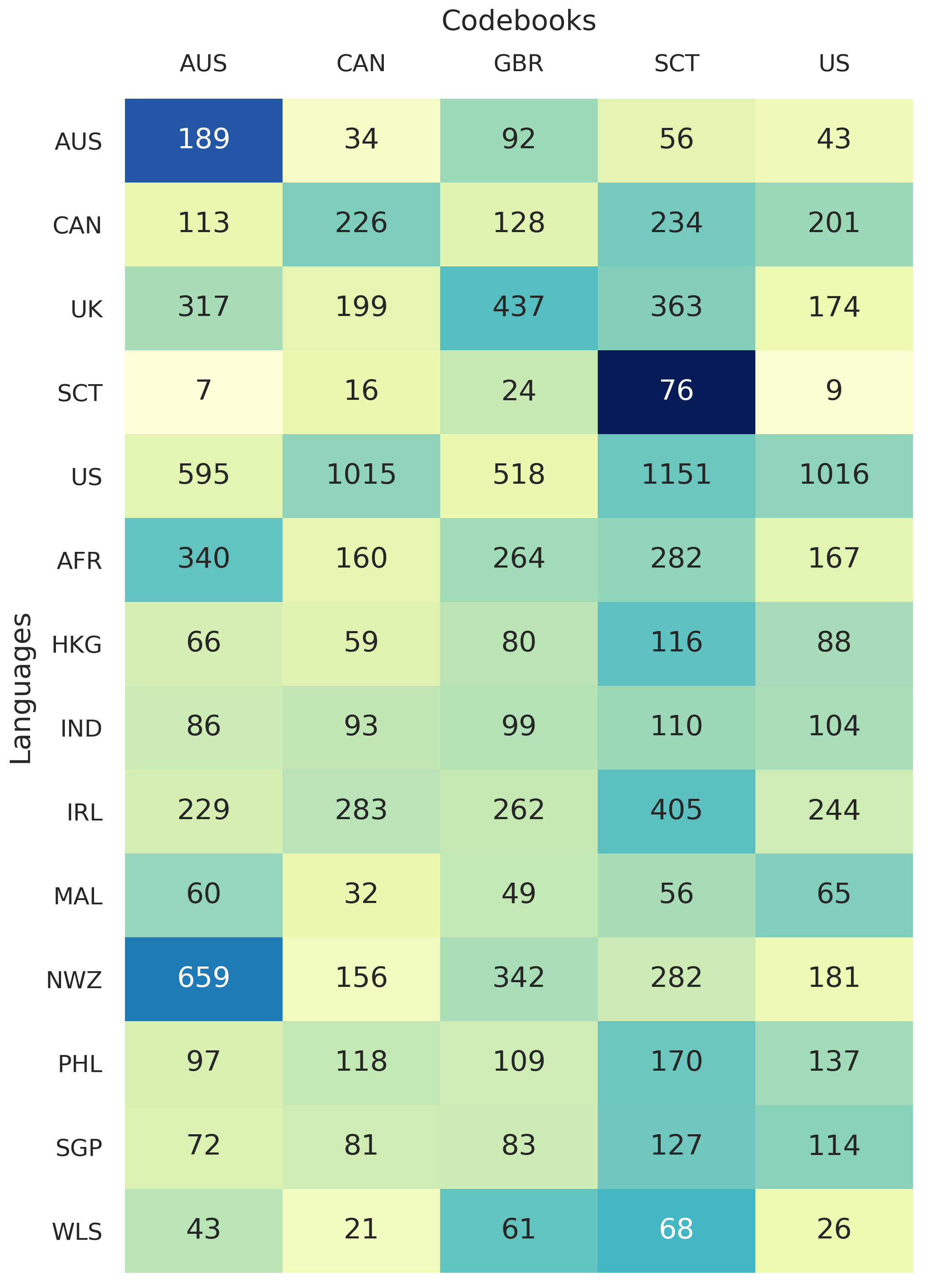}}
\caption{Heatmap showing which codebooks are chosen during inference across seen and unseen accents. For example, the third cell in the first row shows that $92$ out of $413$ Australian-accented utterances used the codebook belonging to England during decoding.}
\label{fig:heatmap}
\vspace{-0.2cm}
\end{figure}

% \begin{figure*}[h]
%     \centering
%     \begin{minipage}[b]{.48\textwidth}
%     \centering
%     \vspace{-0.2cm}
%     \includegraphics[scale=0.47]{average_entropy.png}
%     \caption{Progression of the entropy of the probability distribution across the seen accents averaged over all test samples}
%     \label{fig:entropy}
%   \vspace{-0.2cm}
    
%     \end{minipage}
%     \begin{minipage}[b]{.02\textwidth}
%     \vspace{-0.2cm}
%     \hspace{0.3cm}
%   \vspace{-0.2cm}
    
%     \end{minipage}
%     \begin{minipage}[b]{.48\textwidth}
%     \centering
%     \includegraphics[scale=0.47]{single_example_entropy.png}
%     \vspace{-0.2cm}
%     \caption{Progression of the probability of a seen accented sample appearing in the beam for a single test example}
%     \label{fig:probability}
%   \vspace{-0.2cm}
    
%     \end{minipage}
%   \centering
%   \vspace{-0.2cm}
% \end{figure*}

\paragraph{Codebook Utilization:} Figure~\ref{fig:heatmap} is a heatmap showing which accent codebook is used by the joint beam-search algorithm in generating the best ASR hypothesis for the test utterances. Across seen accents, we see a  diagonal dominance, indicating that seen accents show a preference for their respective codebooks. This effect is especially strong for Australia, England and Scotland accents. US and Canada, on the other hand, have examples evenly divided among each other.  Among unseen accents, the Australia-specific codebook is picked up most by New Zealand test utterances. 

\begin{figure}[h]
    \centering
    \includegraphics[scale=0.37]{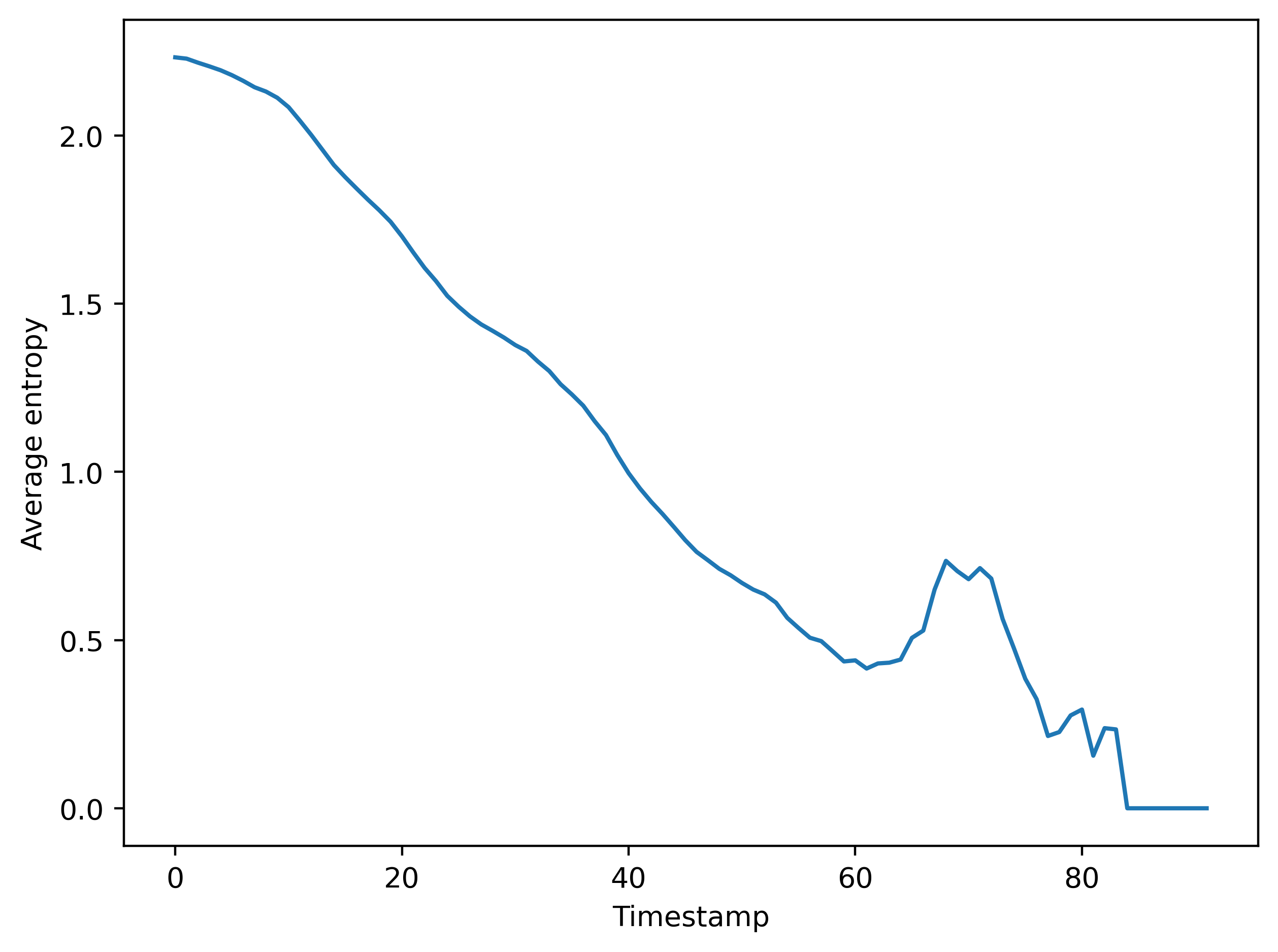}
    %\caption{Progression of the entropy of the probability distribution across the seen accents averaged over all test examples.}
    \caption{Progression of average test entropy of the probability distribution across seen accents.}
    \label{fig:entropy}
  \vspace{-0.5cm}
\end{figure}

\begin{figure}[t!]
    \centering
    \includegraphics[scale=0.47]{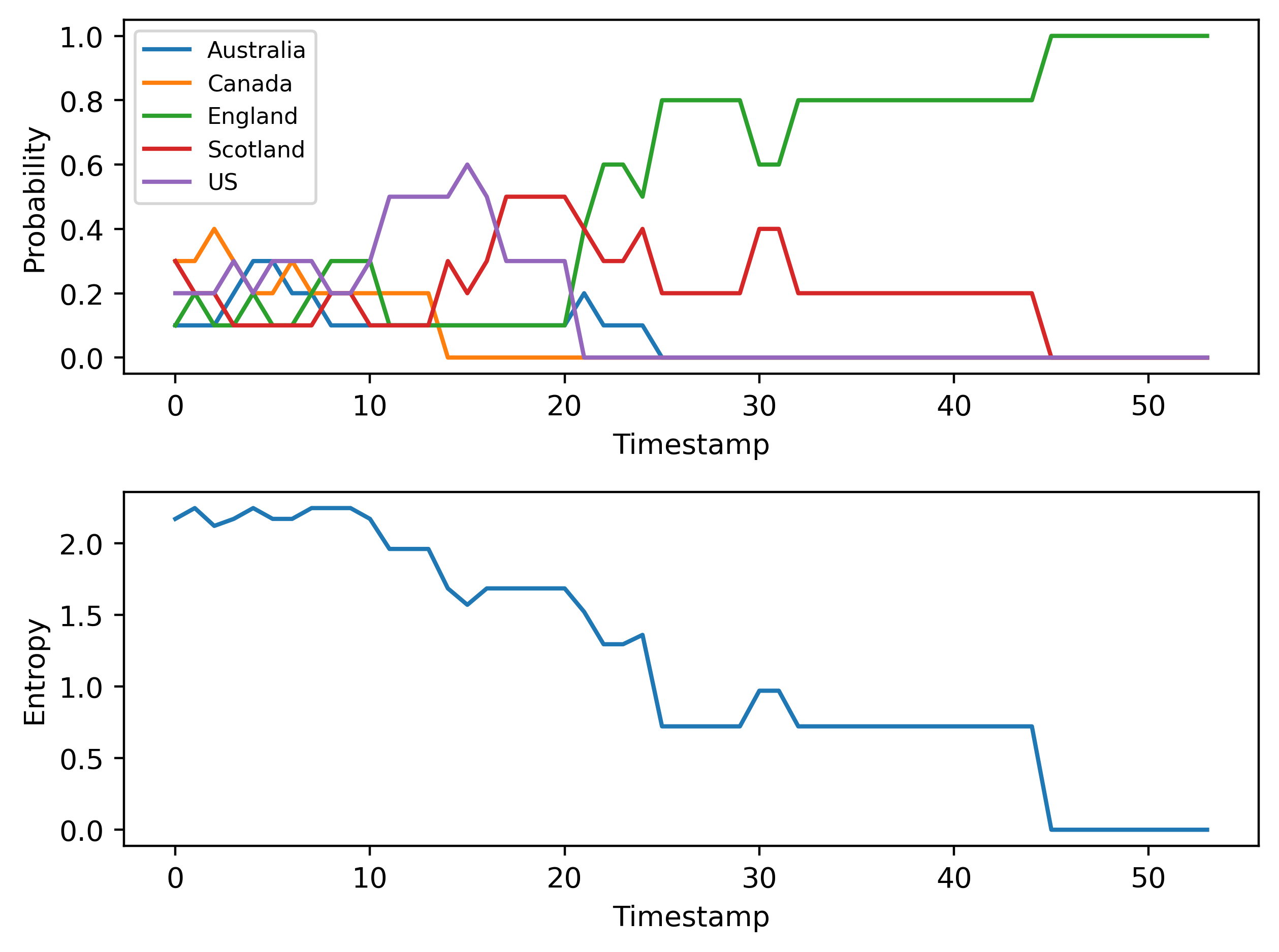}
    \vspace{-0.2cm}
    \caption{Progression of the probability/entropy across seen accents for a single Wales-accented test sample.}
    \label{fig:probability}
  \vspace{-0.4cm}
\end{figure}

\paragraph{Active Accents during Joint Beam-search:} Using joint beam-search decoding, it is possible for samples from different accents to get pruned in early iterations and only one or two dominant accents to be active from the start. To check for this, we compute the distribution of samples in the beam across the five seen accents and plot the average entropy of this distribution across all test instances in Figure~\ref{fig:entropy}. It is clear that four to five seen accents are active until time-step 20, after which certain accents gain more prominence. Figure~\ref{fig:probability} shows both the probabilities across seen accents appearing in the beam for a single Wales-accented test sample, along with the entropy of this distribution. This shows how nearly all accents are active at the start of the utterance, with England becoming the dominant accent towards the end.
\looseness=-1
%This entropy plot shows how (nearly) all accents are in the running until time-step 20, after which it slowly converges to one or two accents.
%possible for the correct accent to be pruned in early beam-search iterations due to low scores. To check the possibility of such a behavior, across decoding time-steps, we show, in Figure~\ref{fig:entropy} the entropy of the probability distribution across seen accents averaged across all test samples and  in Figure~\ref{fig:probability} the probability of a seen accented sample appearing in the beam for a single test example. From the entropy plot, it is clear that at least four or all five seen accents are active until time-step 20, after which certain accents gain more prominence. The probability plot shows how (nearly) all accents are in the running until time-step 20, after which it slowly converges to one or two accents.

\paragraph{Alternatives to Joint Beam-search:} We also explore two alternatives to learning accent labels within the ASR model itself:
%Since the accent labels are absent during inference, our first intuition was to handle this within the end-to-end model. So we experimented with two architectural modifications: 
\begin{enumerate*}[label=\roman*)]
    \item We jointly trained an accent classifier with ASR. During inference, this classifier provides pseudo-accent labels across seen accents that we use to choose the codebook.
    \item We adopted a gating mechanism inspired by~\citet{zhang2021share} that adds a learnable gate to each codebook entry. 
    %As in our current proposal, codebook entries are split across the seen accents. 
    Unlike our current deterministic policy of picking a fixed subset of codebook entries, the learned gates are trained jointly with ASR to pick a designated codebook entry corresponding to the underlying accent of the utterance. During inference, the learned gates determine the codebook entries to be used for each encoder layer.
\end{enumerate*}
Both these techniques performed better than the Conformer baseline but were equivalent in performance to the DAT approach \cite{bobw}. We hypothesize that this could be due to the lack of a strong accent classifier (or a lack of appropriate learning in the gates to capture accent information). Our joint beam-search decoding bypasses this requirement by searching across all seen accents.
\looseness=-1

\paragraph{Why do we see Performance Improvements on Unseen Accents?} For test utterances from unseen accents, our model is designed to choose (seen) accent codebooks that best fit the underlying (unseen) accent. It is somewhat analogous to how humans use familiar accents to tackle unfamiliar ones~\cite{eol,accent_experience}. During inference, our model searches through seen accent codebooks and chooses entries that are most like the unseen accents in the test instances.

\section{Conclusion}
\label{sec:concl}

In this work, we propose a new end-to-end technique for accented ASR that uses accent-specific codebooks and cross-attention to achieve significant performance improvements on seen and unseen accents at test time. We experiment with the Mozilla Common Voice corpus and show detailed ablations over our design choices. We also empirically analyze whether our codebooks encode information relevant to accents. The effective use of codebooks for accents opens up future avenues to encode non-semantic cues in speech that affect ASR performance, such as types of noise, dialects, emotion styles of speech, etc.  
%We modify a Conformer-based ASR system to attend to additional accent-specific information and learn accent-aware representations. This additional information comes through codebooks that are learned during training.  We conduct many tests over our proposed technique using Common Voice corpus and describe in detail the effectiveness of our architecture through many ablation studies and analyses. In addition, we also achieve significant WER reductions on out-of-domain examples.  

\section*{Acknowledgements}
The second and third authors gratefully acknowledge financial support from a SERB Core Research Grant, Department of Science and Technology, Govt of India on accented speech processing.

\section{Limitations}
\label{sec:lim}

We identify a few key limitations of our proposed approach:
\begin{itemize}
    \item The codebook size is a hyperparameter that needs to be finetuned for each task.
    \item We currently employ accent-specific codebooks, one for each accent. This does not scale very well and also does not enable sharing of codebook entries across accent codebooks. Instead, we could use a single (large) codebook and use learnable gates to pick a subset of codebook entries corresponding to the underlying accent of the utterance. 
    \item Our proposed joint beam-search leads to a $16\%$ increase in computation time at inference. This can be made more efficient as part of future work.    \item Our joint beam-search allows for each utterance at test-time to commit to a single seen accent. However, parts of an utterance might benefit from one seen accent, while other parts of the same utterance might benefit from a different seen accent. Such a mix-and-match across seen accents is currently not part of our approach. Accommodating for such effects might improve our model further. 
    %(Gated codebook entries described in Section~\ref{sec:analysis} is one way towards a mix-and-match strategy across seen accents for a single test utterance.)
\end{itemize}
%Although our approach shows a significant improvement over other methodologies, using an oracle experiment, we find that a further 3\% absolute improvement is achievable. This suggests a clear scope for improvement, especially on the accent identifier front. We also find that the same setup does not necessarily perform well on all datasets. For example, a bigger dataset benefits better from a larger codebook. Thus, a  hyper-parameter tuning step needs to be done for a specific dataset before arriving at the best model. Further, as seen in Table~\ref{table:res_inference}, the proposed CA approach leads to a $16$\% increase in the computation time at inference. 

\bibliography{emnlp2023}
\bibliographystyle{acl_natbib}

\appendix

\begin{table*}[t!]
\centering
\resizebox{\linewidth}{!}{
    \begin{tabular}{ l | c | c | l  }
        \hline
        \hline
        
        \multicolumn{1}{c|}{\textbf{\small{Accent}}} & \textbf{\small{Ground Truth}} & \multicolumn{1}{c|}{\textbf{\small{Experiment}}}  & \textbf{\footnotesize{Sentence}} \\

        \hline

        \multirow{3}{*}{ Australia } & \multirow{3}{5cm}{\centering \texttt{where is your father} } & \small{Base} & \texttt{where is \colorbox{red!40}{youfada}} \\
       & & \small{DAT} & \texttt{where is \colorbox{red!40}{youfada}} \\
       &  & \small{CA} & \texttt{where is \colorbox{green!40}{your father}} \\
       
       \hline

        \multirow{3}{*}{ Canada } & \multirow{3}{5cm}{\centering \texttt{putting a pool under this floor was a great idea} } & \small{Base} & \texttt{\colorbox{red!20}{put}\colorbox{red!40}{it} a pool \colorbox{red!40}{into} this floor was a \colorbox{red!40}{greek of}\colorbox{green!40}{idea}} \\
       & & \small{DAT} & \texttt{\colorbox{red!20}{put}\colorbox{red!40}{his} pool \colorbox{green!40}{under} this floor was a \colorbox{green!40}{great}\colorbox{red!40}{of itea}} \\
       &  & \small{CA} & \texttt{\colorbox{green!40}{putting} a pool \colorbox{green!40}{under} this floor was a \colorbox{green!40}{great idea}}  \\

       \hline

       \multirow{3}{*}{ England } & \multirow{3}{5cm}{\centering \texttt{will you breakfast with me} } & \small{Base} & \texttt{will you \colorbox{red!20}{break}\colorbox{red!40}{for swimming}} \\
       & & \small{DAT} & \texttt{will you \colorbox{red!40}{bright for study}} \\
       &  & \small{CA} & \texttt{will you \colorbox{green!40}{breakfast with me}}  \\

       \hline

       \multirow{3}{*}{ Scotland } & \multirow{3}{5cm}{\centering \texttt{elsa knitted furiously} } & \small{Base} & \texttt{elsa knitted \colorbox{red!40}{futiously}} \\
       & & \small{DAT} & \texttt{elsa knitted \colorbox{red!40}{fudiously}} \\
       &  & \small{CA} & \texttt{elsa knitted \colorbox{green!40}{furiously}} \\

       \hline

       \multirow{3}{*}{ US } & \multirow{3}{5cm}{\centering \texttt{how long since weve seen each other} } & \small{Base} & \texttt{\colorbox{red!40}{a}\colorbox{green!40}{long since weve} seen each other} \\
       & & \small{DAT} & \texttt{\colorbox{red!40}{our longsons would} seen each other} \\
       &  & \small{CA} & \texttt{\colorbox{green!40}{how long since weve} seen each other} \\

       \hline

       \multirow{3}{*}{ Africa } & \multirow{3}{5cm}{\centering \texttt{this made them even richer} } & \small{Base} & \texttt{this \colorbox{red!40}{might} them even \colorbox{green!40}{richer}} \\
       & & \small{DAT} & \texttt{this \colorbox{red!40}{might} them even \colorbox{red!40}{richard}} \\
       &  & \small{CA} & \texttt{this \colorbox{green!40}{made} them even \colorbox{green!40}{richer}} \\

       \hline

       \multirow{3}{*}{ Hongkong } & \multirow{3}{5cm}{\centering \texttt{he won a worldwide reputation in his special field} } & \small{Base} & \texttt{he won a \colorbox{red!20}{world}\colorbox{red!40}{white we potason} in his special field} \\
       & & \small{DAT} & \texttt{he won a \colorbox{red!20}{world}\colorbox{red!40}{white repetition} in his special field} \\
       &  & \small{CA} & \texttt{he won a \colorbox{green!40}{worldwide reputation} in his special field} \\

       \hline

       \multirow{3}{*}{ India } & \multirow{3}{5cm}{\centering \texttt{can you play nineties music from paul kelly} } & \small{Base} & \texttt{\colorbox{red!40}{canuble nadis} music from \colorbox{red!40}{policy}} \\
       & & \small{DAT} & \texttt{\colorbox{red!40}{canuple}\colorbox{red!20}{ninety}\colorbox{red!40}{is} music from \colorbox{red!40}{policy}} \\
       &  & \small{CA} & \texttt{\colorbox{green!40}{can you play nineties} music from \colorbox{red!30}{pole}\colorbox{green!40}{kelly}} \\

       \hline

       \multirow{3}{*}{ Ireland } & \multirow{3}{4.8cm}{\centering \texttt{this award is given in three different categories} } & \small{Base} & \texttt{this award is given in \colorbox{red!40}{the dream}\colorbox{green!40}{different categories}} \\
       & & \small{DAT} & \texttt{this award is given in \colorbox{red!40}{the tree giffins categords}} \\
       &  & \small{CA} & \texttt{this award is given in \colorbox{green!40}{three}\colorbox{red!40}{gifting}\colorbox{green!40}{categories}} \\

       \hline

       \multirow{3}{*}{ Malaysia } & \multirow{3}{5cm}{\centering \texttt{it just entered my mind at that moment isaac said} } & \small{Base} & \texttt{\colorbox{red!40}{they} just \colorbox{red!20}{enter} my mind at that moment \colorbox{red!40}{eyes accid}} \\
       & & \small{DAT} & \texttt{\colorbox{green!40}{it} just \colorbox{red!20}{enter} my mind at that moment \colorbox{red!40}{eyes accid}} \\
       &  & \small{CA} & \texttt{\colorbox{green!40}{it} just \colorbox{green!40}{entered} my mind at that moment \colorbox{green!20}{isac}\colorbox{green!40}{said}} \\

       \hline

       \multirow{3}{*}{ Newzealand } & \multirow{3}{5cm}{\centering \texttt{a content delivery network is necessary} } & \small{Base} & \texttt{a content delivery \colorbox{red!40}{mid luke and disincessary}} \\
       & & \small{DAT} & \texttt{a content delivery \colorbox{red!40}{night blue and businecessary}} \\
       &  & \small{CA} & \texttt{a content delivery \colorbox{green!40}{network}\colorbox{red!40}{does}\colorbox{green!40}{necessary}} \\

       \hline

       \multirow{3}{*}{ Singapore } & \multirow{3}{5cm}{\centering \texttt{apple google amazon and facebook are often described as tech giants}} & \small{Base} & \texttt{\colorbox{red!20}{appal} google amazon and \colorbox{green!10}{face book} are often described as \colorbox{red!40}{the} giants} \\
       & & \small{DAT} & \texttt{\colorbox{red!40}{appalo} google amazon and \colorbox{green!20}{face book} are often described as \colorbox{red!40}{tack} giants} \\
       &  & \small{CA} & \texttt{\colorbox{green!40}{apple} google amazon and \colorbox{green!40}{facebook} are often described as \colorbox{green!40}{tech} giants} \\

       \hline

       \multirow{3}{*}{ Philippines } & \multirow{3}{5cm}{\centering \texttt{how many layers of irony are you on} } & \small{Base} & \texttt{how many layers of \colorbox{red!20}{iron}\colorbox{green!40}{are you on}} \\
       & & \small{DAT} & \texttt{how many layers of \colorbox{red!40}{ironea you are}} \\
       &  & \small{CA} & \texttt{how many layers of \colorbox{green!40}{irony}\colorbox{green!40}{are you on}} \\

       \hline

       \multirow{3}{*}{ Wales } & \multirow{3}{5cm}{\centering \texttt{nine rows of soldiers stood in a line} } & \small{Base} & \texttt{\colorbox{red!40}{minros} of soldiers stood in a line} \\
       & & \small{DAT} & \texttt{\colorbox{red!40}{miners} of soldiers stood in a line} \\
       &  & \small{CA} & \texttt{\colorbox{green!40}{nine rows} of soldiers stood in a line} \\

    \hline
    \hline
    \end{tabular}
}

\caption{Comparison of the predictions from the Conformer (labeled as Base), DAT, and our proposed system on a few test utterances. \colorbox{green!40}{$\phantom{x}$} is used to highlight words that are correctly predicted, whereas \colorbox{green!20}{$\phantom{x}$} highlights predictions that are correct but contain minor mistakes such as unwanted spaces. Similarly, \colorbox{red!40}{$\phantom{x}$} denotes incorrectly predicted words, whereas \colorbox{red!20}{$\phantom{x}$} is used to indicate words that are incorrect but are somewhat closer to the underlying transcript such as having similar prefixes.}
\label{table:app_example}
\end{table*}

\section{Dataset Curation}
\label{app:data}
\begin{figure}[ht]
    \centering
	\centerline{\includegraphics[width=0.8\linewidth]{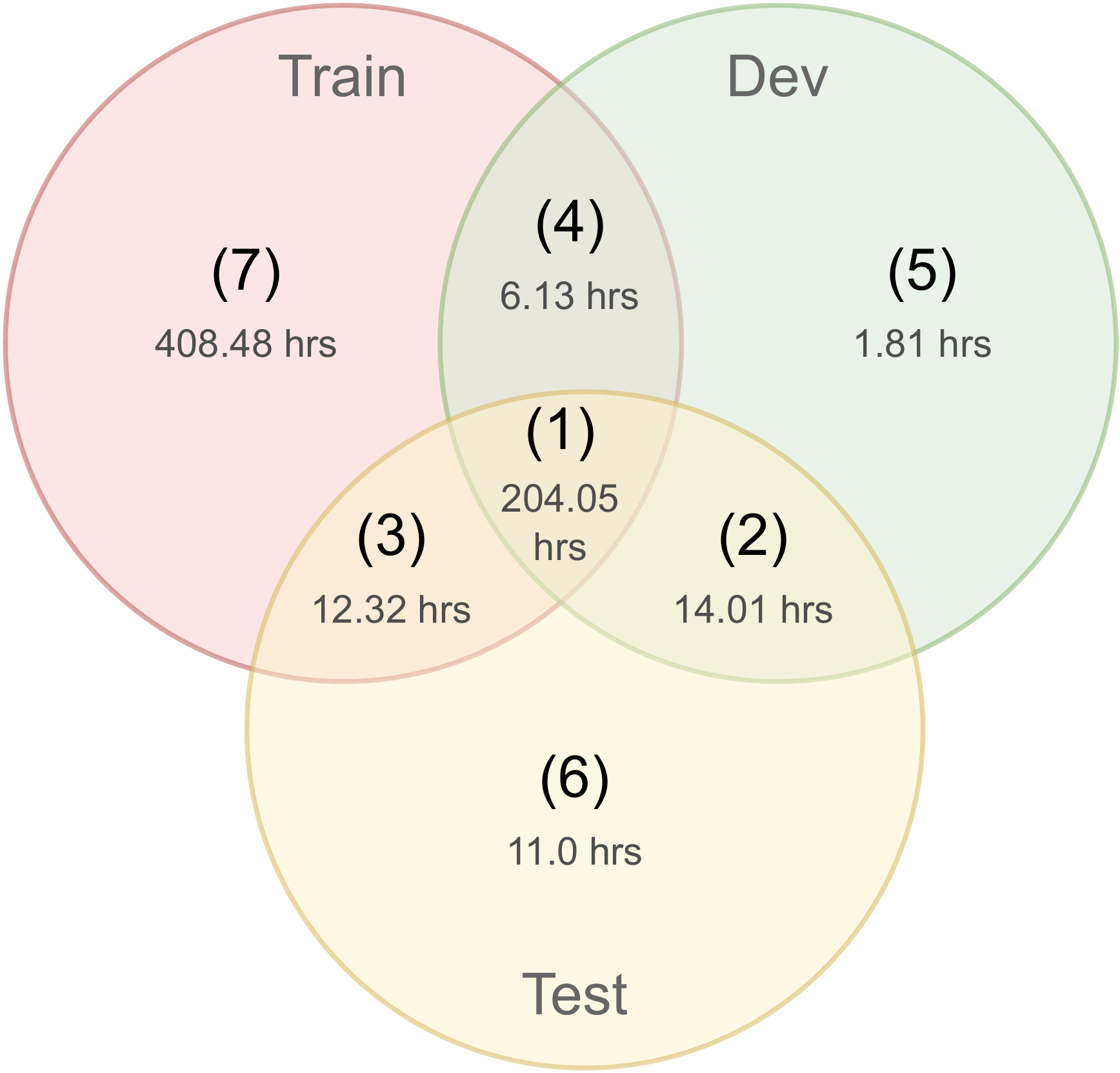}}
	\caption{Illustration of the transcript-wise overlap between the \texttt{train}, \texttt{test}, and \texttt{dev} sets in terms of durations. \textbf{(1)} represents the duration of the group of utterances whose transcripts are present in all three splits.  \textbf{(7)} denotes duration of examples having transcripts found only in the \texttt{train} set.}
	\label{fig:dataset_create}
\end{figure}
%
% \todo{Clearly explain the 204 hours in \(1\): \(1\) is the collection of all utterances that have the their transcript present across all the three sets. 204 hours is the combined duration of that group. These utterances are divided amongst the three sets in some proportion \( with most going for train \). This venn diagram was mostly to show the transcriptwise overlaps between the three sets in terms of durations. }
To build \textsc{mcv\_accent-600}, we group the examples from \mcv into seven buckets while preserving speaker disjointedness across the \texttt{train}, \texttt{dev}, and \texttt{test} sets. 

The buckets are visualized in Figure \ref{fig:dataset_create}; bucket (4) refers to utterances that have exactly the same transcript but different speakers appearing across the train and dev sets. We wanted to include some transcript overlap across all combinations of \texttt{train}, \texttt{test}, and \texttt{dev} splits, since the model could learn accent information from samples with the same transcripts and different underlying accents. We note that a majority of the dev and test samples are disjoint in both speakers and transcripts from the training set for a true evaluation that does not benefit from having seen the same transcripts during training. 

%Since it is essential for the model to be trained on examples with same transcripts across accents so as to get an understanding of accent, the underlying idea is to introduce transcript overlaps across all combinations of \texttt{train}, \texttt{test}, and \texttt{dev} splits.

To create such a split, we loop over all the accents, and for every seen accent $a_{seen}$, we first filter out examples with transcripts that have been previously dealt with and then split the remaining unique transcripts from $a_{seen}$ into seven buckets. For every bucket $b$, transcripts from $b$ are further divided into $n$ groups where $n$ is the number of benefactors for that bucket. As an example, for bucket $(1)$ the value of $n$ is 3. Let $x^i$ be an utterance spoken by speaker $s^j$, which is put into the \texttt{train} set by bucket $(1)$. Then to maintain speaker disjointedness, we put all utterances spoken by $s^j$ into the \texttt{train} set. The transcripts of these utterances are ignored while processing the remaining buckets and accents. For unseen accent $a_{unseen}$, all the examples with transcripts that are not yet processed are put into bucket $(6)$. 

The \texttt{dev} and \texttt{test} sets built this way contain around $68$ and $130$ hours of data. We further randomly sample $25\%$ and $15\%$ from these sets, respectively. To generate \mcv-100 split, we randomly sample $14\%$ from the \mcv-600 set.

\section{Additional Results}
\label{app:results}

Table~\ref{table:app_res_100} shows word error rates (WERs) comparing all experiments done using \textsc{mcv\_accent-100} dataset. Similarly, Table~\ref{table:app_res} shows word error rates comparing all experiments done using \textsc{mcv\_accent-600} dataset.

\begin{table*}[h]
\centering
\resizebox{\linewidth}{!}{
    \begin{tabular}{ l | c | cc | ccccc | ccccccccccc  }
        \hline
        \hline
        
        \multicolumn{1}{c|}{\multirow{2}{*}{\textbf{\small{Method}}}} & \multicolumn{3}{c|}{\textbf{\footnotesize{Aggregated}}} & \multicolumn{5}{c|}{\textbf{\footnotesize{Seen Accents}}} & \multicolumn{9}{c}{\textbf{\footnotesize{Unseen Accents}}} \\
        
        \cline{2-18}

    & \textbf{\scriptsize{Overall}} & \textbf{\scriptsize{Seen}} & \textbf{\scriptsize{Unseen}} & \textbf{\scriptsize{AUS}} & \textbf{\scriptsize{CAN}} & \textbf{\scriptsize{UK}} &
        \textbf{\scriptsize{SCT}}  & \textbf{\scriptsize{US}}  & 
        
    \textbf{\scriptsize{AFR}} & \textbf{\scriptsize{HKG}} & \textbf{\scriptsize{IND}} & \textbf{\scriptsize{IRL}} & \textbf{\scriptsize{MAL}} & \textbf{\scriptsize{NWZ}} & \textbf{\scriptsize{PHL}} & \textbf{\scriptsize{SGP}} & \textbf{\scriptsize{WLS}}  \\
        
        \hline
        
    \small{Transformer} & \footnotesize{22.68} &  \footnotesize{17.34} &  \footnotesize{28.01} &  \footnotesize{18.11} & \footnotesize{17.81} & \footnotesize{19.73} & \footnotesize{18.50} & \footnotesize{16.31} & \footnotesize{25.86} & \footnotesize{31.97} & \footnotesize{35.39} & \footnotesize{25.25} & \footnotesize{36.25} & \footnotesize{23.83} & \footnotesize{31.50} & \footnotesize{38.78} & \footnotesize{21.04} \\
    
    \small{Conformer (Base)} & \footnotesize{18.87} &  \footnotesize{14.05} &  \footnotesize{23.67} & \footnotesize{13.82} & \footnotesize{\textbf{15.02}} & \footnotesize{15.74} & \footnotesize{\textbf{13.36}} & \footnotesize{13.30} & \footnotesize{21.47} & \footnotesize{27.18} & \footnotesize{29.39} & \footnotesize{21.38} & \footnotesize{32.20} & \footnotesize{19.86} & \footnotesize{26.13} & \footnotesize{34.69} & \footnotesize{17.88}  \\
    
    \small{I-vector sum} & \footnotesize{18.87} &  \footnotesize{14.15} & \footnotesize{23.58} & \footnotesize{13.88} & \footnotesize{15.02} & \footnotesize{16.07} & \footnotesize{14.62} & \footnotesize{13.31} & \footnotesize{21.66} & \footnotesize{27.18} & \footnotesize{29.53} & \footnotesize{21.18} & \footnotesize{\textbf{31.72}} & \footnotesize{19.33} & \footnotesize{27.22} & \footnotesize{\textbf{33.81}} & \footnotesize{17.98}  \\
    
    \small{Base + Classifier} & \footnotesize{18.91} &  \footnotesize{14.12} & \footnotesize{23.69} & \footnotesize{14.73} & \footnotesize{15.10} & \footnotesize{16.08} & \footnotesize{13.72} & \footnotesize{13.19} & \footnotesize{21.83} & \footnotesize{28.13} & \footnotesize{\textbf{29.15}} & \footnotesize{21.46} & \footnotesize{32.97} & \footnotesize{19.38} & \footnotesize{26.51} & \footnotesize{34.24} & \footnotesize{18.08}\\
    
    \small{DAT} & \footnotesize{\textbf{18.70}} &  \footnotesize{\textbf{14.00}} & \footnotesize{\textbf{23.38}} & \footnotesize{\textbf{13.30}} & \footnotesize{15.30} & \footnotesize{\textbf{15.72}} & \footnotesize{15.52} & \footnotesize{\textbf{13.15}} & \footnotesize{\textbf{21.15}} & \footnotesize{\textbf{26.95}} & \footnotesize{29.53} & \footnotesize{\textbf{21.15}} & \footnotesize{32.16} & \footnotesize{\textbf{19.22}} & \footnotesize{\textbf{26.03}} & \footnotesize{34.43} & \footnotesize{\textbf{17.93}}  \\
    
    \hline

    \small{$\textsc{CA}_{L\in (1,\ldots,12)}(P=25)$} & \footnotesize{18.33} &  \footnotesize{13.76} & \footnotesize{22.89} & \footnotesize{13.05} & \footnotesize{14.90} & \footnotesize{15.33} & \footnotesize{10.92} & \footnotesize{13.12} & \footnotesize{20.82} & \footnotesize{26.41} & \footnotesize{29.29} & \footnotesize{20.70} & \footnotesize{31.55} & \footnotesize{18.56} & \footnotesize{26.10} & \footnotesize{33.30} & \cellcolor{green!20} \footnotesize{\textbf{16.58}}  \\

    \small{$\textsc{CA}_{L\in (1,\ldots,12)}(P=50)$} & \cellcolor{green!20} \footnotesize{\textbf{18.22}} &  \cellcolor{green!20} \footnotesize{\textbf{13.57}} & \footnotesize{22.86} & \cellcolor{green!20} \footnotesize{\textbf{11.54}} & \cellcolor{green!20} \footnotesize{\textbf{14.81}} & \footnotesize{14.91} & \cellcolor{green!20} \footnotesize{\textbf{9.66}} & \footnotesize{13.15} & \footnotesize{20.95} & \footnotesize{25.66} & \footnotesize{29.15} & \footnotesize{20.72} & \footnotesize{30.87} & \cellcolor{green!20} \footnotesize{\textbf{18.47}} & \cellcolor{green!20} \footnotesize{\textbf{25.81}} & \footnotesize{33.68} & \footnotesize{17.92}  \\

    \small{$\textsc{CA}_{L\in (1,\ldots,12)}(P=100)$} & \footnotesize{18.36} &  \footnotesize{13.85} & \footnotesize{22.86} &  \footnotesize{12.89} & \footnotesize{14.91} & \footnotesize{15.46} & \footnotesize{10.92} & \footnotesize{13.24} & \footnotesize{20.77} & \footnotesize{25.89} & \footnotesize{28.87} & \cellcolor{green!20} \footnotesize{\textbf{20.41}} & \footnotesize{32.44} & \footnotesize{18.76} & \footnotesize{26.24} & \cellcolor{green!20} \footnotesize{\textbf{32.93}} & \footnotesize{17.77} \\

    \small{$\texttt{CA}_{L\in (1,\ldots,12)}(P=200)$} & \footnotesize{18.41} &  \footnotesize{13.69} & \footnotesize{23.12} & \footnotesize{13.00} & \footnotesize{14.81} & \footnotesize{15.06} & \footnotesize{11.10} & \footnotesize{13.12} & \footnotesize{21.57} & \footnotesize{26.78} & \cellcolor{green!20} \footnotesize{\textbf{28.30}} & \footnotesize{20.93} & \footnotesize{30.95} & \footnotesize{18.97} & \footnotesize{25.97} & \footnotesize{33.17} & \footnotesize{17.88} \\

    \small{$\textsc{CA}_{L\in (1,\ldots,12)}(P=500)$} & \footnotesize{18.39} &  \footnotesize{13.68} & \footnotesize{23.09} &  \footnotesize{12.04} & \footnotesize{14.93} & \footnotesize{15.40} & \footnotesize{11.10} & \cellcolor{green!20} \footnotesize{\textbf{13.05}} & \footnotesize{21.22} & \footnotesize{26.52} & \footnotesize{28.77} & \footnotesize{20.57} & \footnotesize{33.13} & \footnotesize{18.78} & \footnotesize{25.97} & \footnotesize{33.81} & \footnotesize{18.39} \\

    \small{$\textsc{CA}_{L\in (1,\ldots,4)}(P=50)$} & \footnotesize{18.30} &  \footnotesize{13.95} & \cellcolor{green!20} \footnotesize{\textbf{22.64}} & \footnotesize{13.22} & \footnotesize{15.53} & \footnotesize{15.49} & \footnotesize{10.74} & \footnotesize{13.24} & \footnotesize{20.79} & \footnotesize{26.06} & \footnotesize{28.58} & \footnotesize{20.52} & \footnotesize{31.43} & \footnotesize{17.87} & \footnotesize{26.27} & \footnotesize{32.98} & \footnotesize{17.72}   \\

    \small{$\textsc{CA}_{L\in (1,\ldots,8)}(P=50)$} & \footnotesize{18.31} &  \footnotesize{13.86} & \footnotesize{22.75} & \footnotesize{13.14} & \footnotesize{15.07} & \footnotesize{15.76} & \footnotesize{10.56} & \footnotesize{13.12} & \cellcolor{green!20} \footnotesize{\textbf{20.50}} & \cellcolor{green!20} \footnotesize{\textbf{25.52}} & \footnotesize{28.70} & \footnotesize{21.03} & \cellcolor{green!20} \footnotesize{\textbf{30.70}} & \footnotesize{18.53} & \footnotesize{25.59} & \footnotesize{33.10} & \footnotesize{18.03} \\

    \small{$\textsc{CA}_{L\in (9,\ldots,12)}(P=50)$} & \footnotesize{18.92} &  \footnotesize{14.24} & \footnotesize{23.59} &  \footnotesize{13.30} & \footnotesize{15.36} & \footnotesize{15.53} & \footnotesize{13.27} & \footnotesize{13.67} & \footnotesize{21.50} & \footnotesize{27.18} & \footnotesize{28.89} & \footnotesize{21.61} & \footnotesize{32.52} & \footnotesize{18.98} & \footnotesize{26.64} & \footnotesize{34.43} & \footnotesize{19.59} \\

    \small{$\textsc{CA}_{L\in (5,\ldots,12)}(P=50)$} & \footnotesize{18.45} &  \footnotesize{13.84} & \footnotesize{23.05} & \footnotesize{12.37} & \footnotesize{15.43} & \footnotesize{15.42} & \footnotesize{10.92} & \footnotesize{13.18} & \footnotesize{21.08} & \footnotesize{26.75} & \footnotesize{28.40} & \footnotesize{20.60} & \footnotesize{31.76} & \footnotesize{18.68} & \footnotesize{26.52} & \footnotesize{34.00} & \footnotesize{18.13} \\

    \small{$\textsc{CA}_{L\in (1,\ldots,12)}(P_\text{rand}=50)$} & \footnotesize{18.30} &  \footnotesize{13.65} & \footnotesize{22.95} & \footnotesize{12.34} & \footnotesize{15.19} & \cellcolor{green!20} \footnotesize{\textbf{14.89}} & \footnotesize{11.64} & \footnotesize{13.06} & \footnotesize{20.77} & \footnotesize{25.57} & \footnotesize{29.48} & \footnotesize{20.98} & \footnotesize{31.88} & \footnotesize{18.62} & \footnotesize{25.87} & \footnotesize{33.28} & \footnotesize{17.82} \\
    \hline
    \hline
    \end{tabular}}
    \caption{Comparison of the performance(WER \% ) of all the experiments mentioned for \textsc{mcv\_accent-100}. Numbers in bold denote the best across baselines, and the green highlighting \colorbox{green!20}{$\phantom{x}$} denotes the best across all the experiments. Ties are broken using overall WER. $\textsc{CA}_{L \in (i,\ldots,j)}(P=k)$: Codebook attend - Cross Attention applied at all layers from $i$ to $j$ with $k$ entries per accent codebook. $\textsc{CA}_{L \in (i,\ldots,j)}(P_\text{rand}=k)$: Similar to the previous setup, but with codebooks frozen during training.}
\label{table:app_res_100}
\end{table*}

\begin{table*}[h]
\centering
\resizebox{\linewidth}{!}{
    \begin{tabular}{ l | c | cc | ccccc | ccccccccccc  }
        \hline
        \hline
        
        \multicolumn{1}{c|}{\multirow{2}{*}{\textbf{\small{Method}}}} & \multicolumn{3}{c|}{\textbf{\footnotesize{Aggregated}}} & \multicolumn{5}{c|}{\textbf{\footnotesize{Seen Accents}}} & \multicolumn{9}{c}{\textbf{\footnotesize{Unseen Accents}}} \\
        
        \cline{2-18}

    & \textbf{\scriptsize{Overall}} & \textbf{\scriptsize{Seen}} & \textbf{\scriptsize{Unseen}} & \textbf{\scriptsize{AUS}} & \textbf{\scriptsize{CAN}} & \textbf{\scriptsize{UK}} &
        \textbf{\scriptsize{SCT}}  & \textbf{\scriptsize{US}}  & 
        
    \textbf{\scriptsize{AFR}} & \textbf{\scriptsize{HKG}} & \textbf{\scriptsize{IND}} & \textbf{\scriptsize{IRL}} & \textbf{\scriptsize{MAL}} & \textbf{\scriptsize{NWZ}} & \textbf{\scriptsize{PHL}} & \textbf{\scriptsize{SGP}} & \textbf{\scriptsize{WLS}}  \\
        
        \hline

    \small{Conformer (Base)} & \footnotesize{9.75} & \cellcolor{green!20} \textbf{\footnotesize{6.04}} &  \footnotesize{13.46} & \footnotesize{4.95} & \cellcolor{green!20} \footnotesize{\textbf{6.81}} & \footnotesize{7.15} & \footnotesize{4.42} & \cellcolor{green!20} \footnotesize{\textbf{5.64}} & \footnotesize{11.81} & \footnotesize{16.50} & \footnotesize{14.90} & \textbf{\footnotesize{12.61}} & \footnotesize{20.43} & \footnotesize{10.49} & \textbf{\footnotesize{15.74}} & \footnotesize{21.26} & \cellcolor{green!20} \footnotesize{\textbf{7.98}} \\

    \small{I-vector} & \footnotesize{10.05} &  \footnotesize{6.40} & \footnotesize{13.69} & \footnotesize{4.67} & \footnotesize{7.53} & \footnotesize{7.43} & \footnotesize{4.06} & \footnotesize{6.04} & \footnotesize{12.05} & \footnotesize{17.45} & \footnotesize{14.76} & \footnotesize{13.08} & \footnotesize{20.31} & \footnotesize{10.57} & \footnotesize{15.82} & \textbf{\footnotesize{21.22}} & \footnotesize{8.91} \\

    \small{MTL} & \footnotesize{10.02} &  \footnotesize{6.33} & \footnotesize{13.70} & \footnotesize{5.30} & \footnotesize{7.59} & \footnotesize{7.39} & \textbf{\footnotesize{3.97}} & \footnotesize{5.86} & \footnotesize{12.19} & \footnotesize{16.30} & \textbf{\footnotesize{14.25}} & \footnotesize{13.23} & \footnotesize{19.58} & \footnotesize{10.64} & \footnotesize{16.30} & \footnotesize{21.54} & \footnotesize{8.50} \\
    
    \small{DAT} & \textbf{\footnotesize{9.73}} &  \footnotesize{6.12} & \textbf{\footnotesize{13.33}} & \footnotesize{\textbf{4.56}} & \footnotesize{7.50} & \footnotesize{\textbf{6.87}} & \footnotesize{4.96} & \footnotesize{5.74} & \textbf{\footnotesize{11.76}} & \textbf{\footnotesize{16.19}} & \footnotesize{14.62} & \footnotesize{12.96} & \textbf{\footnotesize{18.97}} & \textbf{\footnotesize{9.91}} & \footnotesize{15.84} & \footnotesize{21.50} & \footnotesize{8.13} \\
    
    \hline

    \small{$\textsc{CA}_{L\in (1,\ldots,12)}(P=50)$} & \footnotesize{9.63} &  \footnotesize{6.22} & \footnotesize{13.03} & \footnotesize{4.67} & \textbf{\footnotesize{7.36}} & \footnotesize{7.11} & \footnotesize{3.07} & \textbf{\footnotesize{5.90}} & \footnotesize{11.63} & \footnotesize{15.64} & \footnotesize{14.06} & \footnotesize{12.48} & \footnotesize{18.97} & \cellcolor{green!20} \textbf{\footnotesize{9.73}} & \footnotesize{15.60} & \footnotesize{21.05} & \footnotesize{8.29}  \\

    \small{$\textsc{CA}_{L\in (1,\ldots,12)}(P=200)$} & \footnotesize{9.59} &  \footnotesize{6.20} & \footnotesize{12.98} & \footnotesize{4.92} & \footnotesize{7.47} & \footnotesize{6.83} & \footnotesize{3.52} & \footnotesize{5.91} & \cellcolor{green!20} \textbf{\footnotesize{11.47}} & \footnotesize{15.96} & \cellcolor{green!20} \textbf{\footnotesize{13.87}} & \footnotesize{12.25} & \footnotesize{19.30} & \footnotesize{9.98} & \cellcolor{green!20} \textbf{\footnotesize{15.17}} & \footnotesize{20.90} & \textbf{\footnotesize{8.08}}  \\

    \small{$\textsc{CA}_{L\in (1,\ldots,12)}(P=500)$} & \cellcolor{green!20} \footnotesize{\textbf{9.55}} &  \textbf{\footnotesize{6.19}} & \cellcolor{green!20} \textbf{\footnotesize{12.92}} & \cellcolor{green!20} \textbf{\footnotesize{4.40}} & \footnotesize{7.60} & \cellcolor{green!20} \textbf{\footnotesize{6.67}} & \cellcolor{green!20} \textbf{\footnotesize{2.62}} & \footnotesize{5.99} & \footnotesize{11.57} & \cellcolor{green!20} \textbf{\footnotesize{15.18}} & \footnotesize{14.13} & \cellcolor{green!20} \textbf{\footnotesize{12.20}} & \cellcolor{green!20} \textbf{\footnotesize{17.84}} & \footnotesize{10.00} & \footnotesize{15.34} & \cellcolor{green!20} \textbf{\footnotesize{20.71}} & \footnotesize{8.39} \\
    
    \hline
    \hline
    \end{tabular}
}
\caption{Comparison of the performance(WER \% ) of all the experiments mentioned for \textsc{mcv\_accent-600}. We follow the same notation as in Table~\ref{table:app_res_100}.}
\label{table:app_res}
\end{table*}

\section{Comparison of Predictions}
\label{app:examples}

Table~\ref{table:app_example} highlights a few examples where our proposed system performs significantly better than the Conformer baseline and DAT systems.

\end{document}